\documentclass{article} %
\usepackage[table]{xcolor}
\usepackage{iclr2025_conference,times}
\usepackage{booktabs}

\usepackage{amsmath,amsfonts,bm}

\def\eqref#1{equation~\ref{#1}}

\def\1{\bm{1}}

\DeclareMathAlphabet{\mathsfit}{\encodingdefault}{\sfdefault}{m}{sl}
\SetMathAlphabet{\mathsfit}{bold}{\encodingdefault}{\sfdefault}{bx}{n}

\usepackage{hyperref}
\definecolor{mycitecolor}{RGB}{0,101,177}
\hypersetup{colorlinks=true, citecolor=mycitecolor, pdfborder={0 0 0}}
\usepackage{url}

\usepackage{amsfonts}
\usepackage{nicefrac}
\usepackage{microtype}
\usepackage{xcolor,colortbl}
\usepackage{arydshln}
\usepackage{color}
\usepackage[normalem]{ulem}
\usepackage{graphicx}
\usepackage{amsmath}
\usepackage{amssymb}
\usepackage{epsfig}
\usepackage{algorithm}  
\usepackage{algpseudocode}
\usepackage{pifont}
\usepackage{multirow,multicol,makecell}
\usepackage{bbm}
\usepackage{mathtools}
\usepackage{diagbox}
\usepackage{enumitem}
\usepackage{soul}
\usepackage{tikz}
\usepackage{bbding}
\usepackage{wrapfig}
\usepackage{colortbl}

\usepackage{xspace}
\usepackage{hhline}
\usepackage{caption}

\definecolor{mygray}{gray}{.9}
\definecolor{mytableblue}{rgb}{0.83, 0.90, 0.94}
\definecolor{mytablegreen}{rgb}{0.90, 0.97, 0.87}
\definecolor{mygreen}{RGB}{84,130,53}

\newcommand{\ie}{\textit{i}.\textit{e}.}
\newcommand{\eg}{\textit{e}.\textit{g}.}

\newcommand{\etc}{\textit{etc}}
\def\Ours{\textbf{NOVA}\xspace}

\title{Autoregressive Video Generation without Vector Quantization}

\author{
Haoge Deng\textsuperscript{$1,\!5\!$}\thanks{Equal Contribution. This work was done when the first three authors were interns at Beijing Academy of Artificial Intelligence.\textsuperscript{\dag}Corresponding Author: \textit{wangxinlong@baai.ac.cn}, \textit{qiyg@bupt.edu.cn}} \,, 
Ting Pan\textsuperscript{$2,\!3,\!5*$}, Haiwen Diao\textsuperscript{$4,\!5*$}, Zhengxiong Luo\textsuperscript{$5*$}, Yufeng Cui\textsuperscript{$5$}, \\
{\hspace{0.25em}\textbf{Huchuan Lu\textsuperscript{$4$}, Shiguang Shan\textsuperscript{$2,\!3$}, Yonggang Qi\textsuperscript{$1\,$\dag},
Xinlong Wang\textsuperscript{$5\,$\dag}}} \\[3pt]
{\textsuperscript{$1$}Beijing University of Posts and Telecommunications} \\
{\textsuperscript{$2$}Key Laboratory of Intelligent Information Processing, ICT, CAS} \\
{\textsuperscript{$3$}University of Chinese Academy of Sciences} \\
{\textsuperscript{$4$}Dalian University of Technology \quad \textsuperscript{$5$}Beijing Academy of Artificial Intelligence}
}

\iclrfinalcopy
\begin{document}

\makeatletter
\let\@oldmaketitle\@maketitle
\renewcommand{\@maketitle}{\@oldmaketitle

\vspace{-1em}
\centering
\includegraphics[width=.99\linewidth]{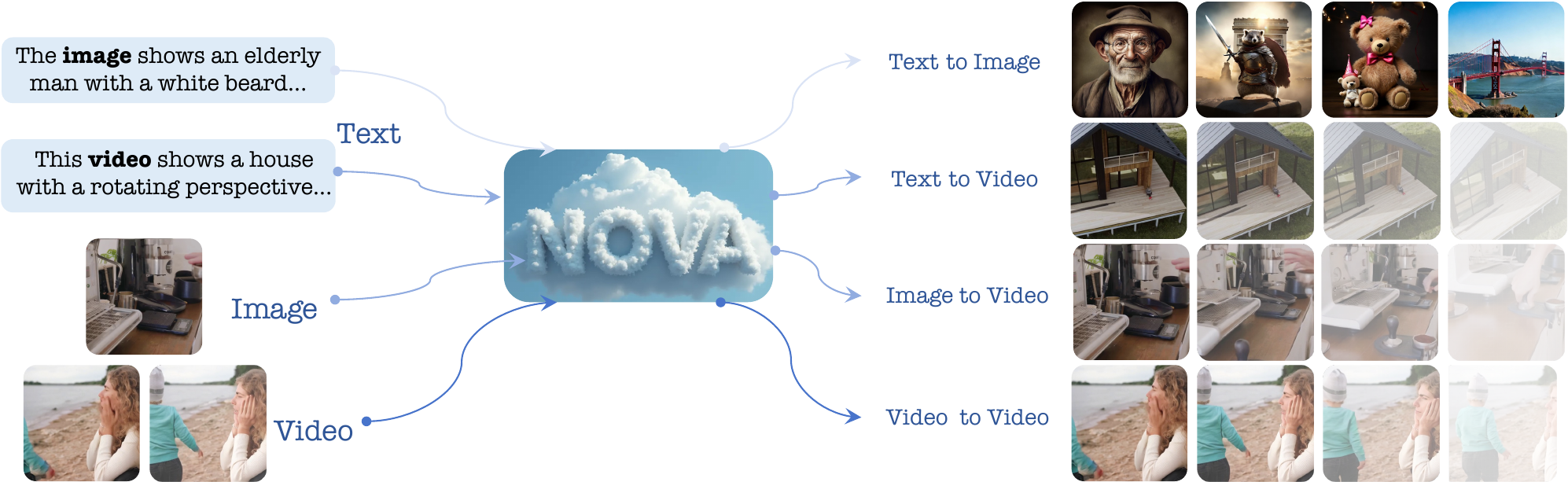} 
\captionsetup{hypcap=false}
\captionof{figure}{
\textbf{NOVA} is a non-quantized autoregressive model for efficient and flexible visual generation.
}\label{fig:teaser}
\vspace{1em}

}
\makeatother

\maketitle

\begin{abstract}
This paper presents a novel approach that enables autoregressive video generation with high efficiency.
We propose to reformulate the video generation problem as a non-quantized autoregressive modeling of temporal \textit{frame-by-frame} prediction and spatial \textit{set-by-set} prediction.
Unlike raster-scan prediction in prior autoregressive models or joint distribution modeling of fixed-length tokens in diffusion models, our approach maintains the causal property of GPT-style models for flexible in-context capabilities, while leveraging bidirectional modeling within individual frames for efficiency. 
With the proposed approach, we train a novel video autoregressive model without vector quantization, termed \Ours.
Our results demonstrate that \Ours surpasses prior autoregressive video models in data efficiency, inference speed, visual fidelity, and video fluency, even with a much smaller model capacity, \ie, 0.6B parameters. 
\Ours also outperforms state-of-the-art image diffusion models in text-to-image generation tasks,  with a significantly lower training cost.
Additionally, \Ours generalizes well across extended video durations and enables diverse zero-shot applications in one unified model. Code and models are publicly available at {\footnotesize \url{https://github.com/baaivision/NOVA}}.

\end{abstract}

\section{Introduction}
Autoregressive large language models (LLMs)~(\cite{TransF:GPT-3, TransF:LLaMA2}) have become a foundational architecture in natural language processing (NLP),
exhibiting emerging capabilities in in-context learning and long-context reasoning.
In autoregressive (AR) vision generation domain, prior approaches~(\cite{MARGEN:DALLE1, MARGEN:COGVIEW1, CAUGEN:PARTI, CAUGEN:VIDEOGPT,  MARGEN:PHENAKI, CAUGEN:VIDEOPOET, EMU3}) typically  transform images or video clips into a discrete-valued token space using vector quantization~(\cite{CAUGEN:VQVAE, CAUGEN:VQGAN}), which are then flattened into sequences for token-by-token prediction.
However, it is challenging for vector-quantized tokenizers to achieve high fidelity and high compression simultaneously.
More tokens are required for a high quality.
Thus, the cost increases substantially with higher image resolutions or longer video sequences.

In contrast, video diffusion models~(\cite{DPM:SORA, DPM:Kling, DPM:SDVIDEODPM}) learn with highly compressed video sequences in a compact continuous latent space.
However, most of them only learn the joint distribution of fixed-length frames, lacking the flexibility to generate videos with varied lengths. More importantly, they do not possess the in-context abilities of autoregressive models, \ie, solving diverse tasks in context with a unified model such as GPT for language. 

In this work, we present \Ours, which addresses the issues above and enables autoregressive video generation with high efficiency. 
We propose to reformulate the video generation problem as a non-quantized autoregressive modeling of temporal \textit{frame-by-frame} prediction and spatial \textit{set-by-set} prediction.
\Ours is inspired by Emu3~(\cite{EMU3}) for autoregressive video and multimodal generation, and MAR~(\cite{MARGEN:MAR}) for non-quantized autoregressive image generation, which utilizes non-quantized vectors as visual tokens and performs set-by-set autoregressive prediction.
While both are non-quantized autoregressive approaches, it is non-trivial at all from MAR to \Ours: \textbf{1)} \Ours solves the challenges including efficiency, scalability, and mask schedule when learning more complex text-to-image generation instead of class-to-image generation.
\textbf{2)} \Ours first predicts temporal frames sequentially and then predicts spatial sets within each frame. \Ours is the first to enable a non-quantized autoregressive model for video generation.

Specifically, \Ours predicts each frame in a casual order temporally, and predicts each token set in a random order spatially. In this way, text-to-video generation can be regarded as a fundamental task that implicitly and comprehensively encompasses various generation tasks (See Figure~\ref{fig:teaser}), including text-to-image, image-to-video, text\&image-to-video, \etc. 
With non-quantized tokenizers and a flexible autoregressive framework, \Ours 
simultaneously takes advantage of \textbf{1}) high-fidelity and compact visual compression for low cost in training and inference, and \textbf{2)} in-context abilities for integrating multiple visual generation tasks in a unified model.

For text-to-video generation, \Ours surpasses autoregressive counterparts in data efficiency, inference speed, and video fluency, while matching the performance of diffusion models of similar scale, \eg, achieving a VBench (\cite{SETUP:VBENCH}) score of 80.1 with a processing speed of 2.75 FPS\footnote{The 2.75 FPS is measured on a single NVIDIA A100-40G GPU using a batch size of 24.}, trained in only 342 GPU days on A100-40G.
For text-to-image generation, NOVA achieves a GenEval (\cite{ghosh2024geneval}) score of 0.75, surpassing previous diffusion models with notably lower training cost, \eg, only 127 GPU days for training this state-of-the-art 0.6B model.
Additionally, \Ours also demonstrates strong zero-shot generalization across various contexts. 
We believe that \Ours paves the way for next-generation video generation, offering possibilities for real-time and infinite video generation, beyond Sora-like video diffusion models.

\section{Related Works}

\subsection{Diffusion Models for Visual Generation}

Diffusion models~(\cite{DPM:DDPM,DPM:SMLM}) have made significant advances in visual generation, including text-to-image tasks~(\cite{DPM:SD3,DPM:DALLE3,DPM:IMAGEN3}) and text-to-video tasks~(\cite{DPM:SORA,VIDEODPM:OPENSORAPLAN,DPM:SDVIDEODPM}). Image diffusion models typically model the joint distribution of fixed-length tokens in pixel~(\cite{DPM:DDPM,DPM:GLIDE,DPM:SIMDPM}) or latent space~(\cite{DPM:LDM,DPM:SD3,DPM:DALLE3,DPM:PIXART}). Besides, video diffusion models further introduce temporal layers to capture relationships between a fixed number of video frames. After training, additional tasks and modalities are added by incorporating extra inference tricks~(\cite{DPM:SDEDIT}), structure moderation~(\cite{DPM:SDVIDEODPM,VIDEODPM:StrDPM,VIDEODPM:MAGICEDIT}), and adapter layers~(\cite{DPM:CONTROLNET,VIDEODPM:ANIMATEDPM}). Although these strategies can be composable, they stand in contrast to the autoregressive approaches~(\cite{CAUGEN:VIDEOPOET,MARGEN:COGVIDEO,TransF:GPT-1,TransF:LLaMA2}), which trains a single model end-to-end for multi-task learning, offering notable context scalability and zero-shot generalizability across diverse application scenarios, especially in extending video generation duration.

\subsection{Autoregressive Models for Visual Generation}

\textbf{Raster-scan Autoregressive Models} are typically implemented on the discrete-valued RGB pixels~(\cite{CAUGEN:VideoPixel,CAUGEN:PARALLEL}) or latent space~(\cite{CAUGEN:VQGAN,CAUGEN:VQVAE}), analogous to their language counterparts~(\cite{TransF:GPT-2,LLM:PALM2}).
Recent studies involve autoregressive transformers to generate token sequences in the raster-scan order for image generation~(\cite{MARGEN:DALLE1, MARGEN:COGVIEW1, MARGEN:COGVIEW2, CAUGEN:PARTI, CAUGEN:LLAMAGEN}), and video generation~(\cite{CAUGEN:VIDEOGPT, CAUGEN:VIDEOPOET, CAUGEN:TRANSFRMSER, CAUGEN:LOONG}).
Specifically, VAR~(\cite{CAUGEN:VAR}) introduces next-scale prediction to progressively process the token-by-token sequence across multiple resolutions, leading to improved image quality.

\textbf{Masked Autoregressive Models} further develop a masked generative models~(\cite{MARGEN:MASKGIT}) to introduce a generalized autoregressive concept. They introduce a bidirectional transformer and predict randomly masked tokens by attending to unmasked conditions. This makes up for the suboptimal modeling and inefficient inference of sequentially line-by-line strategy, which inspires a series of subsequent works in text-to-image~(\cite{MARGEN:MUSE}) and text-to-video generation~(\cite{MARGEN:COGVIDEO,MARGEN:MAGVIT,MARGEN:PHENAKI}).
Particularly, MAR~(\cite{MARGEN:MAR}) decouples discrete tokenizers from autoregressive models and utilizes a diffusion procedure for per-token probability distributions. It is fully validated in the class-to-image field, holding great potential in the text-to-image domain. However, its application to text-to-video generation intuitively requires a masked autoregressive process across entire video frames, challenging multi-context learning and training efficiency. 
In contrast, our NOVA model breaks down video generation into frame-by-frame temporal predictions combined with spatial set-by-set predictions. This allows each frame to act as a meta causal unit, enabling extended video duration and zero-shot generalizability across various contexts. Besides, the subsequent spatial set-of-tokens prediction unlocks the power of bidirectional modeling patterns, enhancing inference efficiency while preserving visual quality and fidelity.

\section{Methodology}
\begin{figure}[t]
\centering 
\includegraphics[width=0.98\linewidth,trim= 0 0 0 0,clip]{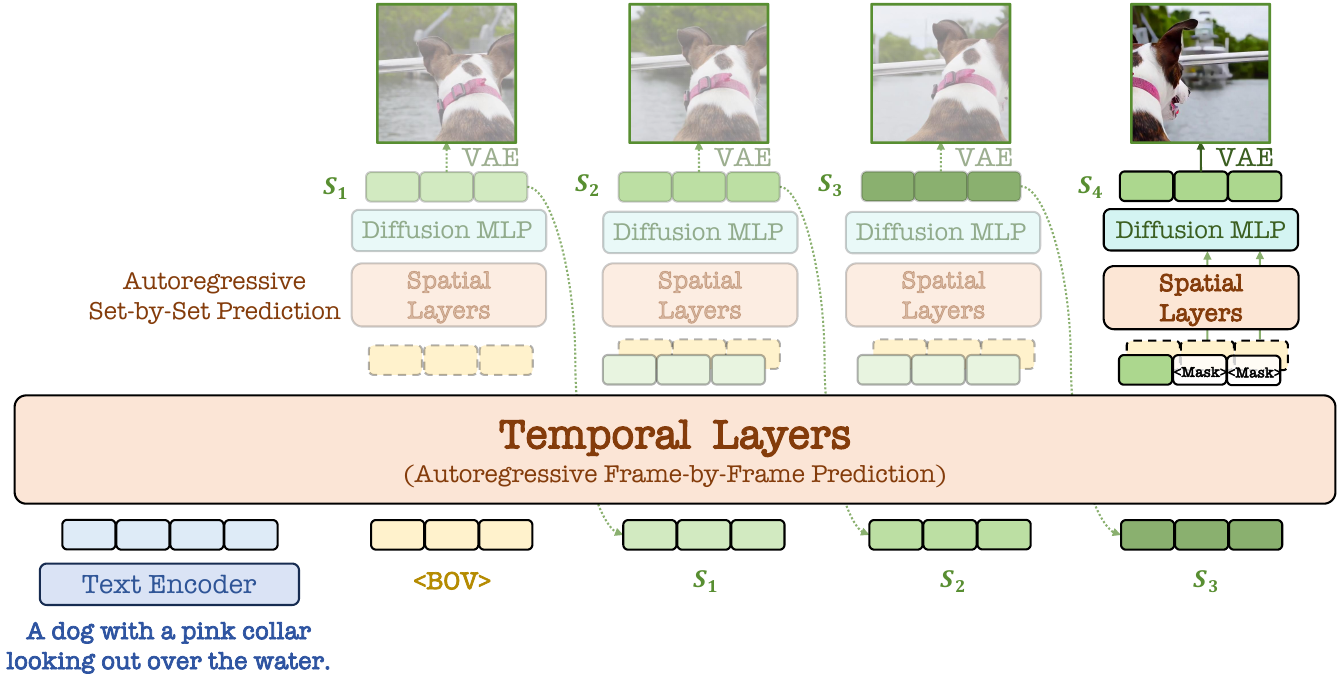} 
\caption{\textbf{NOVA framework and the inference process.} With text inputs, NOVA performs autoregressive generation via temporal frame-by-frame prediction and spatial set-by-set prediction. Finally, we implement diffusion denoising in a continuous-values space.}
\label{fig:framework}
\end{figure}

We first review two categories of autoregressive video generation in Sec.~\ref{sec:rethink}. In Sec.~\ref{sec:TEMAR}-\ref{sec:DOMLOSS}, we introduce framework pipeline and implementation details of our NOVA, illustrated in Figure~\ref{fig:framework}. 

\subsection{Rethinking autoregressive models for video generation}\label{sec:rethink}

\addtolength{\tabcolsep}{-4pt}
\begin{wraptable}{r}{0.40\textwidth}
\vspace{-1.3em}
\caption{Symbology Settings.}
\label{tab:symbology}
\scalebox{0.66}{\Large
\begin{tabular}{l|l}
\toprule
$N,n$ \, & \, The number of all video tokens. \\
$F,f$ \, & \, The number of all video frames. \\
$K,k$ \, & \, The number of sets in an image. \\
\bottomrule 
\end{tabular}}
\vspace{-1.2em}
\end{wraptable}

As mentioned above, we regard text-to-video generation and autoregressive (AR) model as the basic task and means, respectively.
We briefly retrospect related technical background.
There exist two types of AR video generation approaches: \textbf{(1) Token-by-token generation via raster-scan order.} These studies perform causal per-token prediction within video frame sequence~(\cite{CAUGEN:VIDEOPOET}), and decode vision tokens sequentially following the raster scan ordering~(\cite{EMU3}), which is defined as follows:
\begin{equation}
\label{eq:raster-scan-AR}
p\,({C}, {x_{1}},...,{x_{N}})\, = \, \prod_{n}^{N} \, p\,({x_{n}}\,|\,{C},{x_{1}},...,{x_{n-1}})\,,
\end{equation}
where $C$ indicates various condition contexts, \eg, label, text, image, and \etc. Note that $x_{n}$ denotes $n$-th token of $N$ video raster-scale tokens.In contrast, \textbf{(2) Masked set-by-set generation in a random order} treats all tokens within each video frame equally, using a bidirectional transformer decoder for per-set prediction~(\cite{MARGEN:MAGVIT}). However, this generalized autoregressive (AR) model is trained using synchronous modeling on large, fixed-length video frames, which can lead to poor scalability in context and issues with coherence over longer video durations. Hence, NOVA proposes a novel solution by decoupling per-set generation within a single video frame from the per-frame prediction across the entire video sequence. This allows NOVA to better handle both temporal causality and spatial relationships, providing a more flexible and scalable AR framework.

\subsection{Temporal autoregressive modeling via Frame-by-Frame Prediction} \label{sec:TEMAR}

\begin{figure}[t]
\centering
\includegraphics[width=\linewidth]{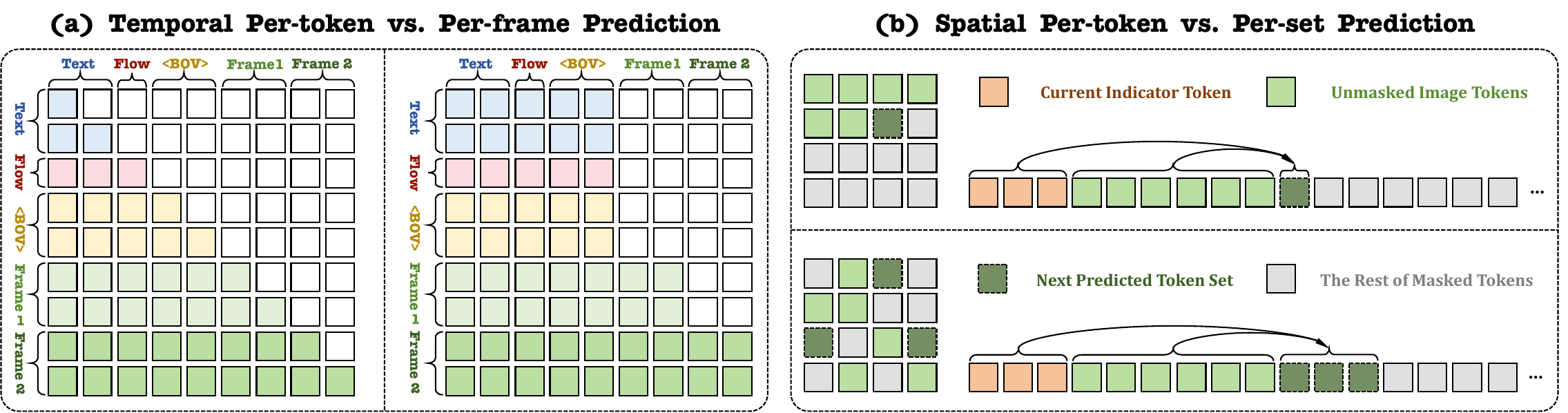}
\caption{Overview of our block-wise temporal and spatial generalized autoregressive attention. Different from per-token generation, NOVA regressively predicts each frame in a casual order across the temporal scale, and predicts each token set in a random order across the spatial scale.}
\label{fig:attention}
\end{figure}

Inspired by ~(\cite{DPM:LUMINANEXT}), we use a pre-trained language model~(\cite{LLM:PHI2}) to encode text prompts to features. To better control video dynamics, we use OpenCV (cv2)~(\cite{VIDEODPM:CV2}) to compute the optical flow of sampled video frames. The average flow magnitude is used as a motion score and integrated with the prompt.
Besides, we employ open-source 3D variational autoencoder (VAE)~(\cite{VIDEODPM:OPENSORAPLAN}) with a temporal stride of 4 and a spatial stride of 8 to encode the video frames to the latent space. We add an additional learnable patch embedding layer with a spatial stride of 4 to align channels of latent video to the subsequent transformer. 
Notably, next-token prediction in early AR models seems counter-intuitive for undirected visual patches within a single image and suffers from high latency during inference. In contrast, video frames can naturally be viewed as a causal sequence, with each frame acting as a meta unit for AR generation. Therefore, we implement block-wise causal masking attention depicted in Figure~\ref{fig:attention}(a), ensuring that each frame can only attend to the text prompts, video flow, and its preceding frames, while allowing all current frame tokens to be visible to each other as follows:
\begin{equation}
\label{eq:temporal-AR}
p\,({P}, {m},{B},{S_{1}},...,{S_{F}})\, = \, \prod_{f}^{F}\,p\,({S_{f}}\,|\,{P}, {m},{B},{S_{1}},...,{S_{f-1}})\,,
\end{equation}
where ${P}, {m}$ indicate text prompts and video flow respectively. Here, ${S_{f}}$ denotes the overall tokens of $f$-th video frame, and ${B}$ represent learnable begin-of-video (BOV) embeddings for predicting the initial video frame, the number of which corresponds to the patch number of one single frame. Note that we add 1-D and 2-D sine-cosine embeddings~(\cite{TransF:Transformer}) with video frame features to indicate time and position information respectively, which are convenient for temporal and spatial extrapolation. From~\eqref{eq:temporal-AR}, we can reformulate text-to-image and image-to-video generation as $p\,({S_{1}}\,|\,{P}, {m},{B})$ and $p\,({S_{f}}\,|\, \varnothing, {m},{B},{S_{1}},...,{S_{f-1}})$. 
This generalized causal process can synchronously model the condition contexts for each video frame, greatly enhancing training efficiency, and allowing the kv-cache technology for fast decoding procedure during inference. 

\subsection{Spatial autoregressive modeling via Set-by-Set Prediction}\label{sec:SPAAR}

Inspired by~(\cite{MARGEN:MASKGIT, MARGEN:MAR}), we define each token set with multiple tokens from random directions as a meta causal token in Figure~\ref{fig:attention}(b), facilitating a generalized AR process with efficient parallel decoding. Notably, we tried to utilize the temporal layers' outputs targeting one frame as indicator features to assist the spatial layers, gradually decoding all randomly masked token sets within the corresponding image. However, this approach resulted in image structure collapse and inconsistent video fluency over the increasing number of frames. 
We hypothesize that this occurs because the indicator features from adjacent frames are similar, making it difficult to accurately learn continuous and imperceptible motion changes without explicit modeling. Besides, the indicator features derived from the ground-truth contextual frame during training contribute to weak robustness and stability of spatial AR layers against cumulative inference errors.

To address this issue, we introduce a Scaling and Shift Layer that reformulates cross-frame motion changes by learning relative distribution variations within a unified space, rather than directly modeling the unreferenced distribution of the current frame. 
Notably, we select the BOV-attended output of the temporal layers as the anchor feature set, as it serves as the initial feature set with significantly less noise accumulation than subsequent frame feature sets. Specifically, we first translate the features from current frame set into dimension-wise variance and mean parameters $\gamma$ and $\beta$ via multi-layer perception (MLP). After that, we affine the normalized features from the anchor set into indicator features $S_{f}^{\prime}$ via channel-wise scale and shift operation. Specially, we explicitly set $\gamma=1$ and $\beta=0$ for the first frame. With unmasked token features, we predict randomly masked visual tokens in a set-by-set order through a bidirectional paradigm, which can be formulated as follows:
\begin{equation}
\label{eq:spatial-AR}
p\,({S^{\prime}_{f}},{S_{(f,1)}},...,{S_{(f,K)}})\, = \, \prod_{k}^{K}\,p\,({S_{(f,k)}}\,|\,{S^{\prime}_{f}},{S_{(f,1)}},...,{S_{(f,k-1)}})\,,
\end{equation}
where $S_{f}^{\prime}$ denotes the indicator features for generating $f$-th video frame, and $S_{(f,k)}$ denotes $k$-th token set of $f$-th video frame. We add 2-D sine-cosine embeddings with masked and unmasked tokens to indicate their relative position.
This generalized spatial AR prediction leverages powerful bidirectional patterns within single-image tokens and achieves efficient inference with parallel masked decoding.
\textit{Notably, we incorporate post-norm layers before the residual connections in both temporal and spatial AR layers.} Our empirical findings show that this design effectively addresses architectural and optimization challenges that previously hindered stable training in generalized video generation.

\subsection{Diffusion Procedure Denoising for Per-token Prediction}\label{sec:DOMLOSS}
During training, we import \textit{diffusion loss}~(\cite{MARGEN:MAR}) to estimate per-token probability in a continuous-valued space. For example, we define one ground-truth token as ${x}_{n}$ and corresponding NOVA's output as ${z}_{n}$. The loss function can be formulated as a denoising criterion:
\begin{equation}
\label{eq:diffusion_loss}
\mathcal{L}({x}_{n}\,|\,{z}_{n})=
\mathbb{E}_{\varepsilon, t}\left[\left\|\epsilon-{\epsilon}_{\theta}\left({x}_{n}^{t} \mid t, {z}_{n}\right)\right\|^{2}\right]\, .
\end{equation}
Here $\epsilon$ is a Gaussian vector sampled from $\mathcal{N}(\mathbf{0}, \mathbf{I})$, and noisy data ${x}_{n}^{t} = \sqrt{\bar{\alpha}_{t}} {x_{n}}+\sqrt{1-\bar{\alpha}_{t}} \epsilon$, where $\bar{\alpha}_{t}$ is a noise schedule~(\cite{DPM:IDDPM}) indexed by a time step $t$. The noise estimator $\epsilon_{\theta}$ is multiple MLP blocks parameterized by $\theta$. The notation $\epsilon_{\theta}({x}_{n}^{t}\, |\, t, {z}_{n})$ means that this network takes ${x}_{n}^{t}$ as the input, and is conditional on both ${t}$ and ${z}_{n}$. We follow~(\cite{MARGEN:MAR}) to sample ${t}$ by 4 times during training for each image.

During inference, we sample ${x}_{n}^{T}$ from a random Gaussian noise $\mathcal{N}(\mathbf{0}, \mathbf{I})$ and denoise it step-by-step by sequentially sampling ${x}_{n}^{T}$ to ${x}_{n}^{0}$ via ${x}_{n}^{t-1}=\frac{1}{\sqrt{\alpha_{t}}}\left({x}_{n}^{t}-\frac{1-\alpha_{t}}{\sqrt{1-\bar{\alpha}_{t}}} \epsilon_{\theta}\left({x}_{n}^{t} | t, z_{n}\right)\right)+\sigma_{t} \epsilon$, where $\sigma_{t}$ is the noise level at time step $t$, and $\epsilon$ is sampled from the Gaussian distribution $\mathcal{N}(\mathbf{0}, \mathbf{I})$. 

\section{Experiment}

\subsection{Experiment Setup}
\textbf{Datasets.}
We involve several diverse, curated, and high-quality datasets to facilitate the training of our NOVA. For text-to-image training, we initially curate 16M image-text pairs sourced from DataComp (\cite{DATA:DATACOMP}), COYO (\cite{DATA:COYO}), Unsplash (\cite{DATA:Unsplash}), and JourneyDB (\cite{DATA:JOURNEYDB}). To explore the scaling properties of NOVA, we expanded the dataset to approximately 600M image-text pairs by selecting more images that have a minimum aesthetic score of 5.0 from LAION (\cite{Datasets:Laion-5b}), DataComp and COYO. For text-to-video training, we select 19M video-text pairs on a subset (\cite{VIDEODPM:OPENSORAPLAN}) of Panda-70M~(\cite{DATA:PANDA}) and internal video-text pairs. We further collect 1M of high-resolution video-text pairs from Pexels (\cite{DATA:Pexels}) to fine-tune our final video generation model. Following~(\cite{VLM:EVE}), we train a caption engine based on Emu2-17B (\cite{VLM:EMUv2}) model to create high-quality descriptions for our image and video datasets. The maximum text length is set to 256.

\textbf{Architectures.}
We mostly follow~(\cite{MARGEN:MAR}) to build NOVA's spatial AR layer and denoising MLP block, including a layer sequence of LayerNorm~(\cite{SETUP:LN}), AdaLN~(\cite{SETUP:ADLN}), linear layer, SiLU activation~(\cite{SETUP:SILU}), and another linear layer. We configure the temporal encoder, spatial encoder, and decoder with 16 layers each, using a dimension of 768 (0.3B), 1024 (0.6B) or 1536 (1.4B). The denoising MLP consists of 3 blocks with a dimension of 1280. The spatial layers adopt the encoder-decoder architecture of MAR (\cite{MARGEN:MAR}), similar to MAE (\cite{TransF:MAE}). Specifically, the encoder processes the visible patches for reconstruction. The decoder further processes visible and masked patches for generation. To capture the image latent features, we employ a pre-trained and frozen VAE from (\cite{VIDEODPM:OPENSORAPLAN}), which achieves $4\times$ compression in the temporal dimension and $8\times8$ compression in the spatial dimension. We adopt the masking and diffusion schedulers from~(\cite{MARGEN:MAR, DPM:IDDPM}), using a masking ratio between 0.7 and 1.0 during training, and progressively reducing it from 1.0 to 0 following a cosine schedule~(\cite{MARGEN:MUSE}) during inference. In line with common practice~(\cite{DPM:DDPM}), we train with a 1000-step noise schedule but default to 100 steps for inference.

\textbf{Training details.}
NOVA is trained with sixteen A100 (40G) nodes. We utilize the AdamW optimizer~(\cite{SETUP:ADAMW}) ($\beta_{1}=0.9,\beta_{2}=0.95$) with a weight decay of 0.02 and a base learning rate of 1e-4 in all experiments. The peak learning rate is adjusted for different batch sizes during training using the scaling rule~(\cite{SETUP:TRAINONEHPU}) : $\text{lr}=\text{base\_lr} \times \text{batchsize} / 256$. We train text-to-image models from scratch and then load these weights to train text-to-video models.

\textbf{Evaluation.}
We use T2I-CompBench~(\cite{SETUP:T2ICOMBENCH}), GenEval (\cite{ghosh2024geneval}) and DPG-Bench (\cite{hu2024ella} to assess the alignment between the generated images and text condition. We generate image samples for each of the original or rewritten (\cite{EMU3}) text prompts. Each image sample has a resolution of 512$\times$512 or 1024$\times$1024. We use VBench~(\cite{SETUP:VBENCH}) to evaluate the capacity of text-to-video generation across 16 dimensions. For a given text prompt, we randomly generate 5 samples, each with a video size of 33$\times$768$\times$480. We employ classifier-free guidance (\cite{ho2022classifier}) with a value of 7.0 along with 128 autoregressive steps to enhance the quality of the generated images and videos in all evaluation experiments.

\begin{table}[th]
\vspace{-2em}
\caption{\textbf{Text-to-image evaluation on various benchmarks.} The best and second-best results are in \colorbox{mytableblue}{blue} and \colorbox{mytablegreen}{green}. The data is from \cite{SETUP:T2ICOMBENCH},\cite{EMU3} and \cite{SD3}.}
\label{tab:res-t2i}
\vspace{-2mm}
\begin{center}
\resizebox{\linewidth}{!}{
\begin{tabular}{lcc|ccc|ccccccc|c|c}
\toprule
\multirow{3}{*}{Model} & \multicolumn{2}{c}{\textbf{ModelSpec}} & \multicolumn{3}{c}{\textbf{T2I-CompBench}} & \multicolumn{7}{c}{\textbf{GenEval}} & \multicolumn{1}{c}{\textbf{DPG-Bench}} & \multirow{3}{*}{\textbf{A100 days}} \\
\cmidrule(lr){2-3}\cmidrule(lr){4-6}\cmidrule(lr){7-13}\cmidrule(lr){14-14}
 & \#params & \#images & Color & Shape & Texture & Overall & Single & Two & Counting & Colors & Position & ColorAttr & Overall \\
\midrule
\rowcolor{mygray}{\textit{Diffusion models}}&&&&&&&&&&&&&&   \\
PixArt-$\alpha$               & 0.6B & 25M  & 68.86 & 55.82 & 70.44 & 0.48 & 0.98 & 0.50 & 0.44 & 0.80 & 0.08 & 0.07 & 71.11 & 753 \\
SD v1.5                       & 1B   & 2B   & 37.50 & 37.24 & 42.19 & 0.43 & 0.97 & 0.38 & 0.35 & 0.76 & 0.04 & 0.06 & 63.18 & - \\
SD v2.1                       & 1B   & 2B   & 56.94 & 44.95 & 49.82 & 0.50 & 0.98 & 0.37 & 0.44 & 0.85 & 0.07 & 0.17 & -     & - \\
SDXL                          & 2.6B & -    & 63.69 & 54.08 & 56.37 & 0.55 & 0.98 & 0.44 & 0.39 & \colorbox{mytablegreen}{0.85} & 0.15 & 0.23 & 74.65 & - \\
DALL-E2                       & 6.5B & 650M & 57.50 & 54.64 & 63.74 & 0.52 & 0.94 & 0.66 & 0.49 & 0.77 & 0.10 & 0.19 & -     & - \\
DALL-E3                       & -    & -    & 81.10 & \colorbox{mytableblue}{67.50} & \colorbox{mytableblue}{80.70} & 0.67 & 0.96 & 0.87 & 0.47 & 0.83 & 0.43 & 0.45 & \colorbox{mytablegreen}{83.50} & - \\
SD3                           & 2B   & -    & -     & -     & -     & 0.62 & 0.98 & 0.74 & \colorbox{mytableblue}{0.63} & 0.67 & 0.34 & 0.36 & \colorbox{mytableblue}{84.10} & - \\
\rowcolor{mygray}{\textit{Autoregressive models}} &&&&&&&&&&&&&& \\
LlamaGen                      & 0.8B & 60M  & -     & -     & -     & 0.32 & 0.71 & 0.34  & 0.21 & 0.58 & 0.07 & 0.04 & - & - \\
Emu3 (+ Rewriter)             & 8B   & -    & 79.13 & 58.46 & 74.22 & 0.66 & \colorbox{mytablegreen}{0.99} & 0.81  & 0.42 & 0.80 & 0.49 & 0.45 & 81.60 & - \\
\midrule 
\midrule 
NOVA (512$\times$512)         & 0.6B & 16M  & 70.75 & 55.98 & 69.79 & 0.66 & 0.98 & 0.85 & 0.58 & 0.83 & 0.20 & 0.48 & 81.76 & 127 \\
\,\, + Rewriter               & 0.6B & 16M  & \colorbox{mytableblue}{83.02} & \colorbox{mytablegreen}{61.47} & \colorbox{mytablegreen}{75.80} & \colorbox{mytableblue}{0.75} & 0.98 & 0.88 & 0.62 & 0.82 & \colorbox{mytableblue}{0.62} & \colorbox{mytableblue}{0.58}  & - & 127 \\
\,\, + Videos                 & 0.6B & 36M  & 71.80 & 47.86 & 65.31 & 0.55 & 0.98 & 0.56 & 0.48 & 0.75 & 0.15 & 0.41 & 81.77 & 342 \\
\,\, + Videos \& Rewriter     & 0.6B & 36M  & \colorbox{mytablegreen}{81.36} & 59.16 & 72.45 & 0.71 & 0.98 & 0.83 & 0.52 & 0.81 & \colorbox{mytablegreen}{0.58} & 0.51 & - & 342 \\
\midrule 
NOVA (1024$\times$1024)       & 0.3B & 600M & 73.35 & 57.28 & 70.09 & 0.67 & 0.98 & 0.86 & 0.53 & 0.84 & 0.32 & 0.52 & 80.60 & 267 \\
NOVA (1024$\times$1024)       & 0.6B & 600M & 74.72 & 56.99 & 69.50 & 0.69 & 0.98 & \colorbox{mytablegreen}{0.89} & 0.56 & 0.84 & 0.32 & 0.56 & 82.25 & 320 \\
NOVA (1024$\times$1024)       & 1.4B & 600M & 74.30 & 57.14 & 70.00 & \colorbox{mytablegreen}{0.71} & \colorbox{mytableblue}{0.99} & \colorbox{mytableblue}{0.91} & \colorbox{mytablegreen}{0.62} & \colorbox{mytableblue}{0.85} & 0.33 & \colorbox{mytablegreen}{0.56} & 83.01 & 608 \\
\bottomrule
\end{tabular}}
\end{center}
\end{table}

\begin{table}[th]
\caption{\textbf{Text-to-video evaluation on VBench.} We have classified existing video generation methods into different categories for better clarity. The baseline data is sourced from \cite{SETUP:VBENCH}.}
\label{tab:res-t2v}
\vspace{-2mm}
\begin{center}
\setlength{\tabcolsep}{1.5mm}
\renewcommand\arraystretch{1.1} 
\resizebox{\linewidth}{!}{
\begin{tabular}{lccc| ccc cccccccc}
\toprule
Model & \#params & \#videos & latency & \makecell{Total\\Score} & \makecell{Quality\\Score} & \makecell{Semantic\\Score} & \makecell{Aesthetic\\Quality} & \makecell{Object\\Class} & \makecell{Multiple\\Objects} & \makecell{Human\\Action} & \makecell{Spatial\\Relationship} & \makecell{Scene} \\
\midrule
\rowcolor{mygray}{\textit{Closed-source models}} &&&&&&&&&&&& \\
Gen-2 & - & - & - & 80.58 & 82.47 & 73.03 & 66.96 & 90.92 & 55.47 & 89.20 & 66.91 & 48.91 \\
Kling (2024-07) & - & - & - & 81.85 & 83.39 & 75.68 & 61.21 & 87.24 & 68.05 & 93.40 & 73.03 & 50.86 \\
Gen-3 & - & - & - & 82.32 & 84.11 & 75.17 & 63.34 & 87.81 & 53.64 & 96.4 & 65.09 & 54.57 \\
\rowcolor{mygray}\textit{Diffusion models (w/ SD init)} &&&&&&&&&&&& \\
LaVie & 3B & 25M & - & 77.08 & 78.78 & 70.31 &  54.94 & 91.82 & 33.32 & 96.8 & 34.09 & 52.69 \\
Show-1 & 4B & 10M & - & 78.93 & 80.42 & 72.98 & 57.35 & 93.07 & 45.47 & 95.60 & 53.50 & 47.03 \\
AnimateDiff-v2  & 1B & 10M & - & 80.27 & 82.90 & 69.75 & 67.16 & 90.90 &  36.88 & 92.60 & 34.60 & 50.19 \\
VideoCrafter-v2.0 & 2B & 10M & - & 80.44 & 82.20 & 73.42 & 63.13 &  92.55 & 40.66 & 95.00 & 35.86 & 55.29 \\
T2V-Turbo (VC2) & 2B & 10M & - & 81.01 & 82.57 & 74.76 & 63.04 & 93.96 & 54.65 & 95.20 & 38.67 & 55.58 \\
\rowcolor{mygray}{\textit{Diffusion models}} &&&&&&&&&&&& \\
OpenSora-v1.1 & 1B & 10M & 48s & 75.66 & 77.74 & 67.36 & 50.12 & 86.76 & 40.97 & 84.20 & 52.47 & 38.63 \\
OpenSoraPlan-v1.1 & 1B & 4.5M & 60s & 78.00 & 80.91 & 66.38 & 56.85 & 76.30 & 40.35 & 86.80 & 53.11 &  27.17 \\
OpenSora-v1.2 & 1B & 32M & 55s & 79.76 & 81.35 & 73.39 & 56.85 & 82.22 & 51.83 & 91.20 & 68.56 & 42.44 \\
CogVideoX & 2B & 35M & 90s & 80.91 & 82.18 & 75.83 & 60.82 & 83.37 & 62.63 & 98.00 & 69.90 & 51.14 \\
\rowcolor{mygray}\multicolumn{3}{l}{\textit{Autoregressive models}} &&&&&&&&&& \\
CogVideo & 9B & 5.4M & - & 67.01 & 72.06 & 46.83 & 38.18 & 73.4 & 18.11 & 78.20 & 18.24 & 28.24 \\
Emu3 & 8B & - & - & 80.96 & 84.09 & 68.43 & 59.64 & 86.17 & 44.64 & 77.71 &  68.73 & 37.11 \\
\midrule 
NOVA & 0.6B & 20M & 12s & 78.48 & 78.96 & 76.57 & 54.52 & 91.36 & 73.46 & 91.20 & 66.37 & 50.16 \\
\,\, + Rewriter  & 0.6B & 20M & 12s & 80.12 & 80.39 & 79.05 & 59.42 & 92.00 & 77.52 & 95.20 & 77.52 & 54.06 \\              
\bottomrule
\end{tabular}
}\end{center}
\end{table}

\subsection{Main Results}

\textbf{NOVA outperforms existing text-to-image models with superior performance and efficiency.}
In Table~\ref{tab:res-t2i}, we compare NOVA with several recent text-to-image models, including PixArt-$\alpha$~(\cite{DPM:PIXART}), SD v1/v2 (\cite{rombach2022high}, SDXL (\cite{DPM:SDXL}), DALL-E2 (\cite{DPM:DALLE2}), DALL-E3 (\cite{betker2023improving}), SD3 (\cite{SD3}), LlamaGen (\cite{CAUGEN:LLAMAGEN} and Emu3 (\cite{EMU3}). After text-to-image training, NOVA achieves state-of-the-art performance on the GenEval benchmark, especially in generating a specified number of targets. Notably, NOVA also achieves leading results on T2I-CompBench and DPG-Bench, excelling at both the small model scale and data scale (requiring only 16\% training overhead of the best competitor PixArt-$\alpha$). \textit{Last but not least, our text-to-video model outperforms most specialized text-to-image models}, \eg, SD v1/v2, SDXL and DALL-E2. This underscores the robustness and versatility of our model in multi-context scenarios, with text-to-video generation as the fundamental training task.

\begin{figure}[t]
\vspace{-2.5em}\centering
\includegraphics[width=0.98\linewidth]{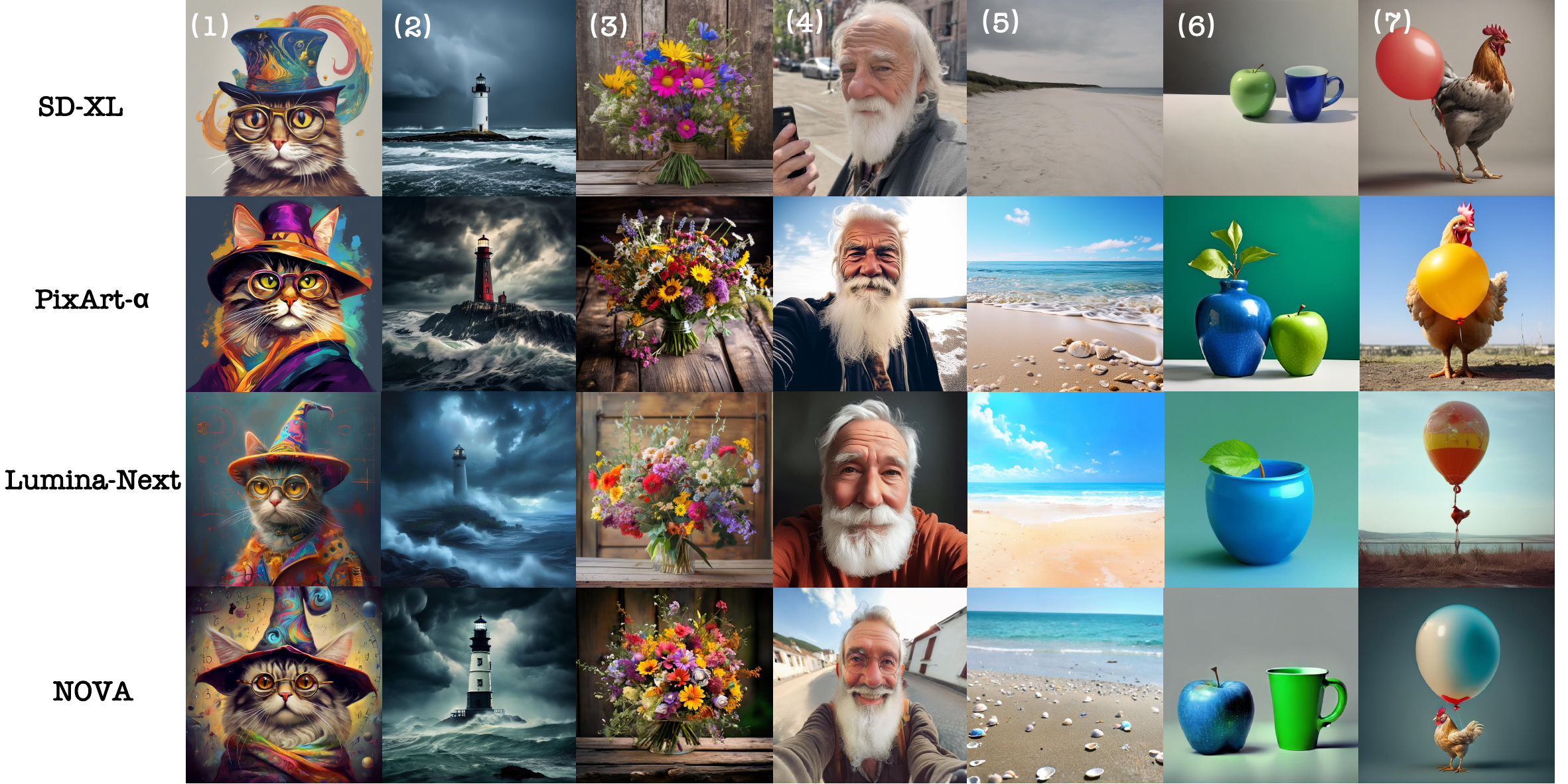} 
\caption{\textbf{Text-to-image generation.} Text prompts from left to right: (1) ``A digital artwork of a cat styled in a whimsical fashion...'', (2) ``A solitary lighthouse standing tall against a backdrop of stormy seas and dark, rolling clouds'', (3) ``A vibrant bouquet of wildflowers on a rustic wooden table'', (4) ``A selfie of an old man with a white beard'', (5) ``A serene, expansive beach with no people'', (6) ``A blue apple and a green cup.'' and (7) ``A chicken on the bottom of a balloon.''}
\label{fig:t2i}
\end{figure}

\textbf{NOVA rivals diffusion text-to-video models and significantly suppresses the AR counterpart.}
We emphasize that the current version of our NOVA is designed to generate videos at 33 frames and can extend video length through the \textit{pre-filling} of recently generated frames. We perform a quantitative analysis comparing NOVA against open-source and proprietary text-to-video models. As shown in Table~\ref{tab:res-t2v}, despite its significantly smaller size (0.6B vs. 9B), NOVA remarkably outperforms CogVideo (\cite{MARGEN:COGVIDEO}) across a variety of text-to-video evaluation metrics. It also matches the latest SOTA model Emu3's~(\cite{EMU3}) performance (80.12 vs. 80.96) with a significantly smaller size (0.6B vs. 8B). Additionally, we compared NOVA with state-of-the-art diffusion models. This includes both the closed-source models (\cite{VIDEODPM:RUNWAYGEN2,DPM:Kling,VIDEODPM:Gen3}), as well as the open-source alternatives (\cite{VIDEODPM:LAVIE,VIDEODPM:Show1,VIDEODPM:AnimateDiff,VIDEODPM:VIDEOCRAFTER2,VIDEODPM:T2VTurbo,VIDEODPM:OPENSORA,VIDEODPM:OPENSORAPLAN,VIDEODPM:COGVIDEOX}).

\subsection{Qualitative Results} 

\begin{figure}[t]
\centering
\includegraphics[width=0.98\linewidth]{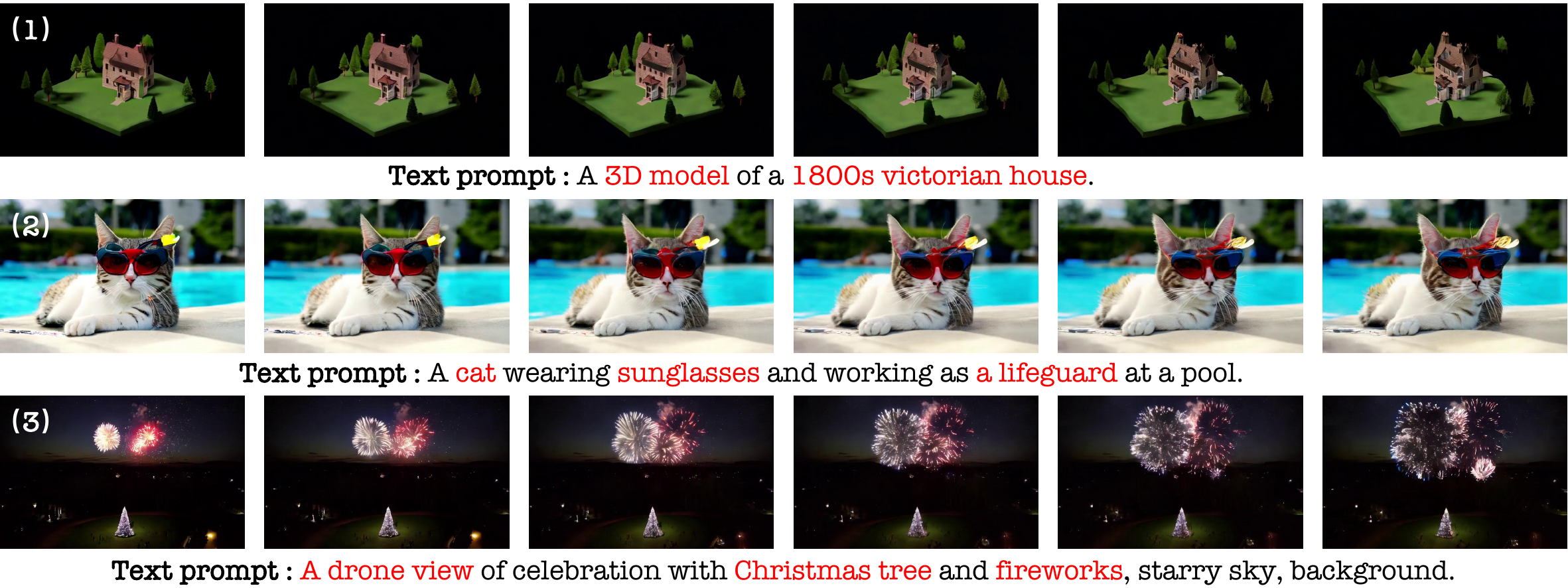} 
\caption{\textbf{Text-to-video generation.} We highlight the keywords in \textcolor{red}{red} color. NOVA follows the text prompts and vividly captures the motion of subjects (i.e., 3D model, cat and fireworks).}
\label{fig:t2v}\vspace{-1em}
\end{figure}

\textbf{High-fidelity image and high-fluency video.} We present a qualitative comparison of current leading image generation methods in Figure~\ref{fig:t2i}. NOVA demonstrates strong visual quality and fidelity across a range of prompts. We present text-to-video visualizations in Figure~\ref{fig:t2v}, which highlight NOVA's ability to capture multi-view perspectives, smooth object motion, and stable scene transitions.

\begin{figure}[t]
\vspace{-2.5em}\centering
\includegraphics[width=0.98\linewidth]{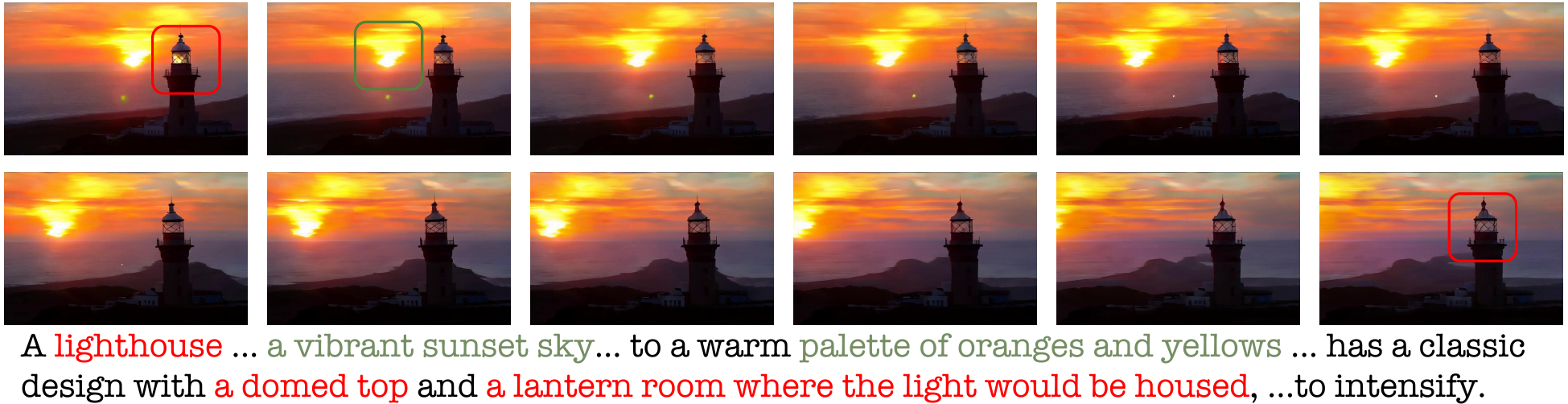}
\caption{\textbf{Zero-shot video extrapolation.} We highlight the subjects in \textcolor{red}{red} and \textcolor{mygreen}{green} respectively. The top images are generated, while the bottom images are extrapolated.}
\label{fig:explo_1}
\end{figure}

\textbf{Zero-shot generalization on video extrapolation.}
By pre-filling generated frames, NOVA can produce videos that surpass the training length. For example, by shifting both the text and BOV embeddings, we generate videos that are up to twice the original length, as shown in Figure~\ref{fig:explo_1}.

\begin{figure}[t]
\vspace{-0.5em}\centering
\includegraphics[width=0.98\linewidth,trim= 0 0 0 0,clip]{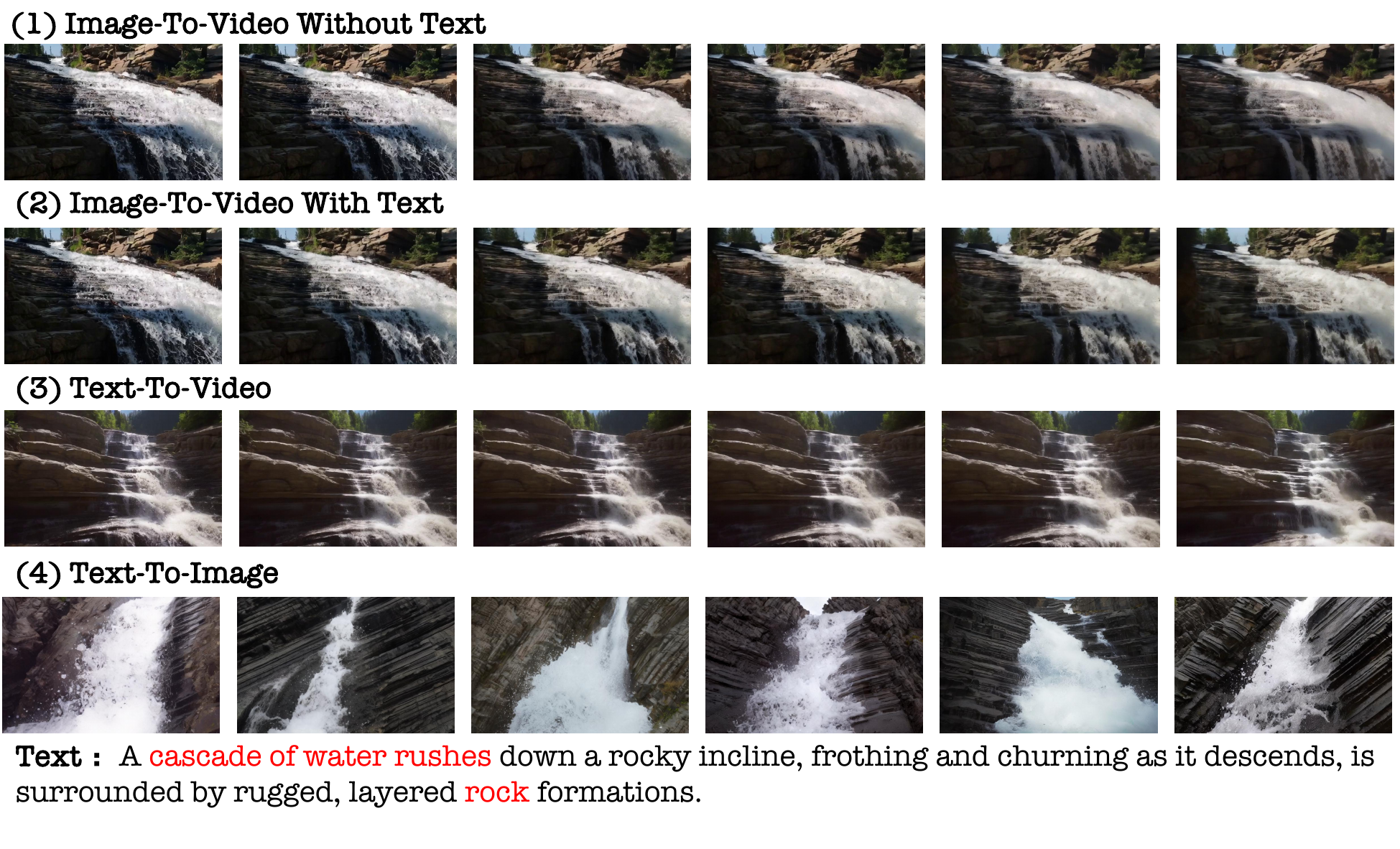}\vspace{-1em}
\caption{\textbf{Zero-shot generalization on multiple contexts.} It is evident that NOVA successfully maintains temporal consistency in objects, both with and without text. Such as ensuring "water continues to flow smoothly." This highlights NOVA's capability for zero-shot multitasking.}
\label{fig:i2v-zero-shot}\vspace{-1em}
\end{figure}

\textbf{Zero-shot generalization on multiple contexts.}
By pre-filling the reference image, NOVA can generate videos from images, either with or without accompanying text. In Figure \ref{fig:i2v-zero-shot}, we provide a qualitative example. We show that NOVA can simulate realistic motions without text prompts. 

\subsection{Ablation Study}

\begin{figure}[t]
\vspace{-3.5em}\centering
\includegraphics[width=0.98\linewidth]{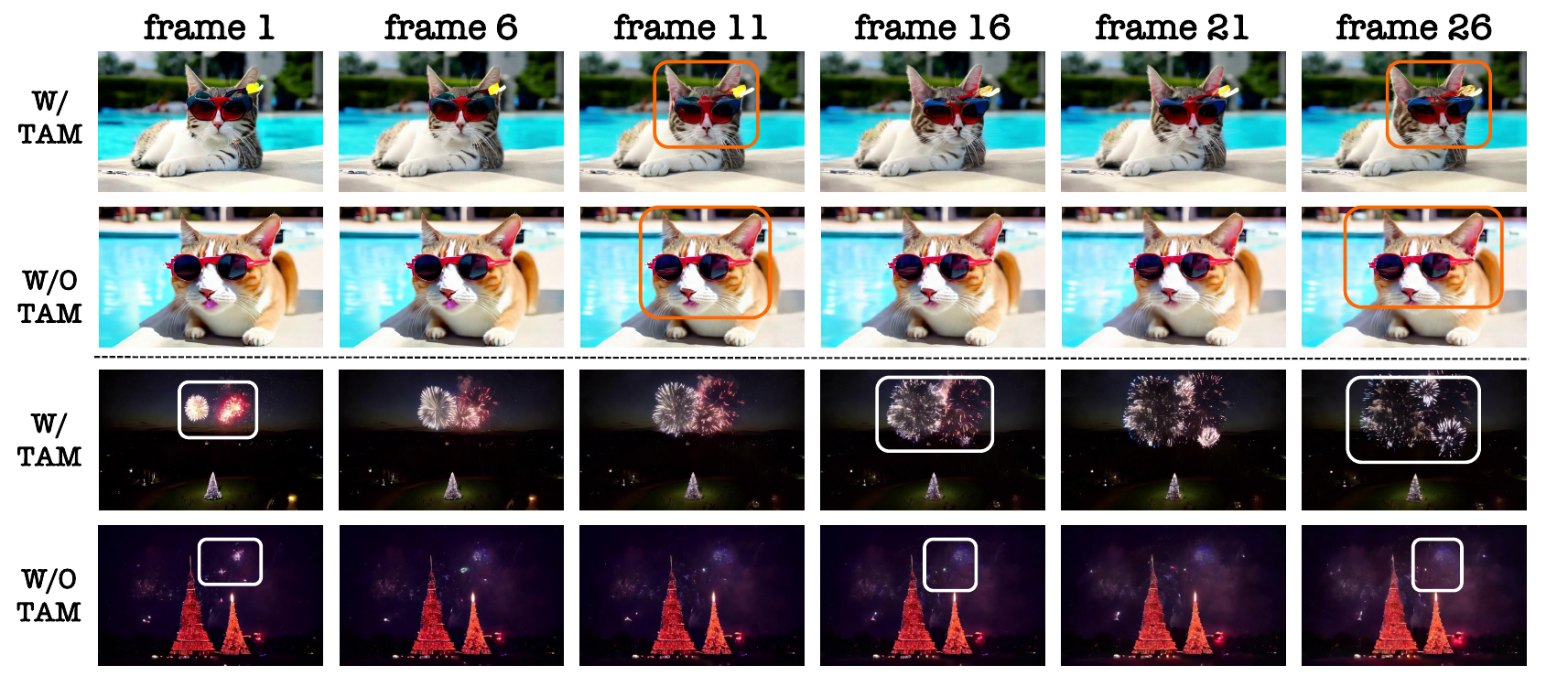} 
\caption{\textbf{Temporal autoregressive modeling (TAM) for video generation.} We highlight the subtle changes in frames generated from the same prompt. Compared to spatial-only autoregressive method, the inclusion of TAM enables NOVA to more accurately capture the dynamics of subject movement.}
\label{fig:mar2d_ablation_visual}
\end{figure}
\begin{figure}[t]
\vspace{-0.5em}\centering
\resizebox{\textwidth}{!}{\begin{tabular}{cc}
\includegraphics[width=0.5\linewidth]{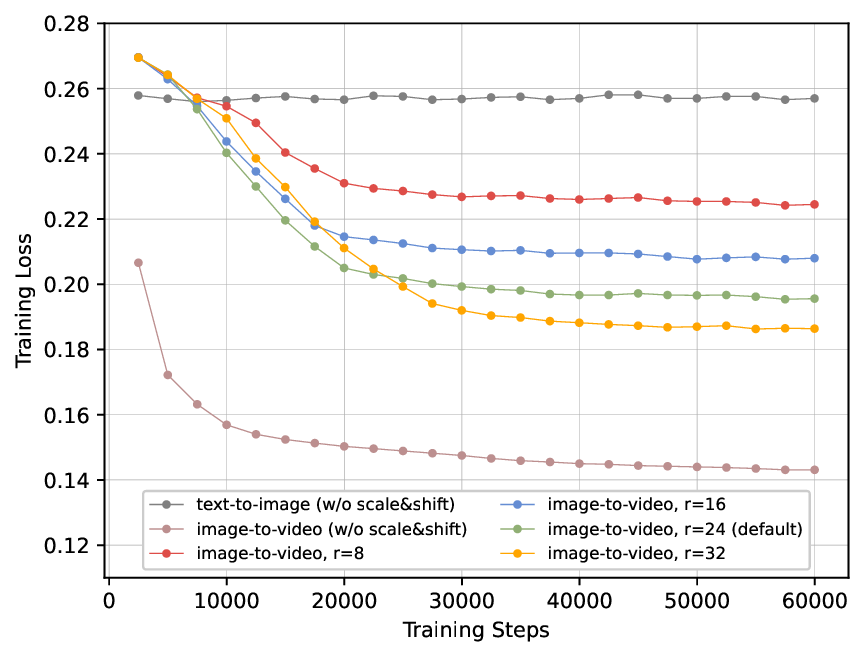} &
\includegraphics[width=0.5\linewidth]{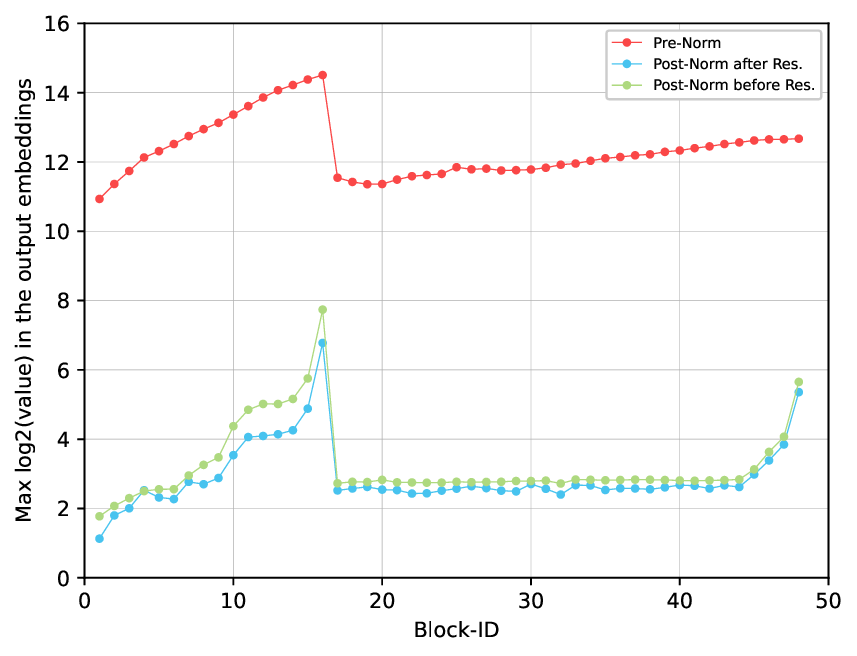} \\
(a) Parameter decomposition for Scaling and Shift layer. & (b) Normalization layer position. \\
\end{tabular}}
\caption{\textbf{Ablation studies on NOVA's architecture components.} We carefully examine the two key stability factors in large-scale video generation training, as illustrated in (a) and (b).}
\label{fig:rank_ablation}\vspace{-1.6em}
\end{figure}

\textbf{Effectiveness of temporal autoregressive modeling.} To highlight the advantages of temporal AR, we have facilitated spatial AR to finish video generation task. We observe less subject movement in videos under the same training iterations (Figure~\ref{fig:mar2d_ablation_visual}). Additionally, in zero-shot generalization across various contexts or video extrapolation, the network output exhibited more artifacts and temporal inconsistencies. Furthermore, this approach is not compatible with kv-cache acceleration during inference, leading to a linear increase in latency with the number of video frames. 

\begin{figure}[th]
\vspace{-0.5em}\centering
\includegraphics[width=0.98\linewidth]{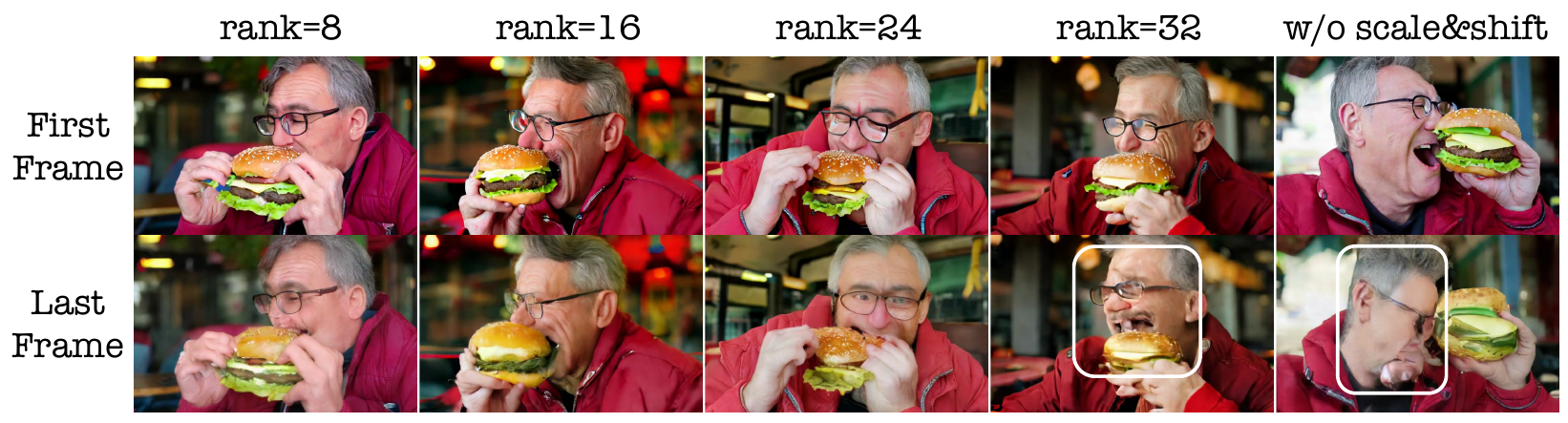} 
\caption{\textbf{Visualization of decomposition ranks in the Scaling and Shift layer.} The first row displays the results of the first frame, while the second row presents the results of the last frame.}
\label{fig:rank_ablation_visual}\vspace{-0.5em}
\end{figure}

\textbf{Effectiveness of Scaling and Shift Layer.}
To capture cross-frame motion changes, we employ a simple yet effective scaling and shifting layer to explicitly model the relative distribution from the BOV-attended feature space. In Figure~\ref{fig:rank_ablation}(a), we demonstrate that this approach significantly reduces the drift between text-to-image and image-to-video generation losses. As we gradually decrease the inner rank of the MLP, the training difficulty increases, leading to a more comprehensive and robust learning process for the network. However, extremely low rank values pose challenges for motion modeling, as they significantly limit the layer's representation capability (Figure~\ref{fig:rank_ablation_visual}). The rank is set to 24 by default in all text-to-video experiments, resulting in more accurate motion predictions.

\textbf{Effectiveness of Post-Norm Layer.}
Training large-scale image and video generation models~(\cite{MARGEN:COGVIEW1, VLM:CHAMELEON}) from scratch often poses significant challenges with mixed precision, which is also observed in other visual recognition methods~(\cite{TransF:SwinTransformerV2}). As shown in Figure~\ref{fig:rank_ablation}(b), the training process with pre-normalization~(\cite{TransF:ViT}) suffers from numerical overflow and variance instability. We attempted various regularization techniques on the residual branch, such as stochastic depth~(\cite{SETUP:STODEPTH}) and residual dropout~(\cite{TransF:Transformer}), but found them to be less effective. Inspired by~(\cite{TransF:SwinTransformerV2}), we introduce post-normalization and empirically discover that it can effectively mitigate the residual accumulation of output embeddings compared to pre-normalization, resulting in a more stable training process.

\section{Conclusion}
In this paper, we present \Ours, a novel autoregressive model designed for both text-to-image and text-to-video generation. \Ours delivers exceptional image quality and video fluency while significantly minimizing training and inference overhead. 
Our key designs include temporal frame-by-frame prediction, spatial set-by-set generation, and continuous-space autoregressive modeling across various contexts. 
Extensive experiments demonstrate that \Ours achieves near-commercial quality in image generation, alongside promising fidelity and fluency in video generation. 
\Ours paves the way for next-generation video generation and world models. It offers valuable insights and possibilities for real-time and infinite video generation, going beyond Sora-like video diffusion models.
As a first step, we will continue scalable experiments with larger models and data scaling to explore NOVA's limits in future work.

\textbf{Acknowledgement.}
We thank Fan Zhang for the valuable advice on Figure 2 of this work, and Xiaotong Li, Jinming Wu, Chengyue Wu, Zhen Li, Quan Sun, Yueze Wang, Jinsheng Wang and Xiaosong Zhang for the insightful discussions. We also acknowledge the support provided by other colleagues at the Beijing Academy of Artificial Intelligence (BAAI) towards this project. This work was supported by the National Key R\&D Program of China (2022ZD0116302), in part by Hainan Provincial Natural Science Foundation of China under No.624LALH008, and in part by the Program for Youth Innovative Research Team of BUPT under No. 2023YQTD02.

\clearpage
\bibliography{iclr2025_conference}

\begin{thebibliography}{96}
\providecommand{\natexlab}[1]{#1}
\providecommand{\url}[1]{\texttt{#1}}
\expandafter\ifx\csname urlstyle\endcsname\relax
  \providecommand{\doi}[1]{doi: #1}\else
  \providecommand{\doi}{doi: \begingroup \urlstyle{rm}\Url}\fi

\bibitem[Anil et~al.(2023)Anil, Dai, Firat, Johnson, Lepikhin, Passos, Shakeri,
  Taropa, Bailey, Chen, et~al.]{LLM:PALM2}
Rohan Anil, Andrew~M Dai, Orhan Firat, Melvin Johnson, Dmitry Lepikhin,
  Alexandre Passos, Siamak Shakeri, Emanuel Taropa, Paige Bailey, Zhifeng Chen,
  et~al.
\newblock Palm 2 technical report.
\newblock \emph{arXiv preprint arXiv:2305.10403}, 2023.

\bibitem[Baldridge et~al.(2024)Baldridge, Bauer, Bhutani, Brichtova, Bunner,
  Chan, Chen, Dieleman, Du, Eaton-Rosen, et~al.]{DPM:IMAGEN3}
Jason Baldridge, Jakob Bauer, Mukul Bhutani, Nicole Brichtova, Andrew Bunner,
  Kelvin Chan, Yichang Chen, Sander Dieleman, Yuqing Du, Zach Eaton-Rosen,
  et~al.
\newblock Imagen 3.
\newblock \emph{arXiv preprint arXiv:2408.07009}, 2024.

\bibitem[Betker et~al.(2023{\natexlab{a}})Betker, Goh, Jing, Brooks, Wang, Li,
  Ouyang, Zhuang, Lee, Guo, et~al.]{DPM:DALLE3}
James Betker, Gabriel Goh, Li~Jing, Tim Brooks, Jianfeng Wang, Linjie Li, Long
  Ouyang, Juntang Zhuang, Joyce Lee, Yufei Guo, et~al.
\newblock Improving image generation with better captions.
\newblock \emph{Computer Science. https://cdn. openai. com/papers/dall-e-3.
  pdf}, 2\penalty0 (3):\penalty0 8, 2023{\natexlab{a}}.

\bibitem[Betker et~al.(2023{\natexlab{b}})Betker, Goh, Jing, Brooks, Wang, Li,
  Ouyang, Zhuang, Lee, Guo, et~al.]{betker2023improving}
James Betker, Gabriel Goh, Li~Jing, Tim Brooks, Jianfeng Wang, Linjie Li, Long
  Ouyang, Juntang Zhuang, Joyce Lee, Yufei Guo, et~al.
\newblock Improving image generation with better captions.
\newblock \emph{Computer Science. https://cdn. openai. com/papers/dall-e-3.
  pdf}, 2\penalty0 (3):\penalty0 8, 2023{\natexlab{b}}.

\bibitem[Blattmann et~al.(2023)Blattmann, Dockhorn, Kulal, Mendelevitch,
  Kilian, Lorenz, Levi, English, Voleti, Letts, et~al.]{DPM:SDVIDEODPM}
Andreas Blattmann, Tim Dockhorn, Sumith Kulal, Daniel Mendelevitch, Maciej
  Kilian, Dominik Lorenz, Yam Levi, Zion English, Vikram Voleti, Adam Letts,
  et~al.
\newblock Stable video diffusion: Scaling latent video diffusion models to
  large datasets.
\newblock \emph{arXiv preprint arXiv:2311.15127}, 2023.

\bibitem[Bradski(2000)]{VIDEODPM:CV2}
G.~Bradski.
\newblock {The OpenCV Library}.
\newblock \emph{Dr. Dobb's Journal of Software Tools}, 2000.

\bibitem[Brooks et~al.(2024)Brooks, Peebles, Holmes, DePue, Guo, Jing, Schnurr,
  Taylor, Luhman, Luhman, et~al.]{DPM:SORA}
Tim Brooks, Bill Peebles, Connor Holmes, Will DePue, Yufei Guo, Li~Jing, David
  Schnurr, Joe Taylor, Troy Luhman, Eric Luhman, et~al.
\newblock Video generation models as world simulators, 2024.
\newblock URL
  \url{https://openai.com/research/video-generation-modelsas-world-simulators}.

\bibitem[Brown et~al.(2020)Brown, Mann, Ryder, Subbiah, Kaplan, Dhariwal,
  Neelakantan, Shyam, Sastry, Askell, Agarwal, Herbert{-}Voss, Krueger,
  Henighan, Child, Ramesh, Ziegler, Wu, Winter, Hesse, Chen, Sigler, Litwin,
  Gray, Chess, Clark, Berner, McCandlish, Radford, Sutskever, and
  Amodei]{TransF:GPT-3}
Tom~B. Brown, Benjamin Mann, Nick Ryder, Melanie Subbiah, Jared Kaplan,
  Prafulla Dhariwal, Arvind Neelakantan, Pranav Shyam, Girish Sastry, Amanda
  Askell, Sandhini Agarwal, Ariel Herbert{-}Voss, Gretchen Krueger, Tom
  Henighan, Rewon Child, Aditya Ramesh, Daniel~M. Ziegler, Jeffrey Wu, Clemens
  Winter, Christopher Hesse, Mark Chen, Eric Sigler, Mateusz Litwin, Scott
  Gray, Benjamin Chess, Jack Clark, Christopher Berner, Sam McCandlish, Alec
  Radford, Ilya Sutskever, and Dario Amodei.
\newblock Language models are few-shot learners.
\newblock In \emph{Advances in Neural Information Processing Systems}, 2020.

\bibitem[Byeon et~al.(2022)Byeon, Park, Kim, Lee, Baek, and Kim]{DATA:COYO}
Minwoo Byeon, Beomhee Park, Haecheon Kim, Sungjun Lee, Woonhyuk Baek, and
  Saehoon Kim.
\newblock Coyo-700m: Image-text pair dataset.
\newblock \url{https://github.com/kakaobrain/coyo-dataset}, 2022.

\bibitem[ChameleonTeam(2024)]{VLM:CHAMELEON}
ChameleonTeam.
\newblock Chameleon: Mixed-modal early-fusion foundation models.
\newblock \emph{arXiv preprint arXiv:2405.09818}, 2024.

\bibitem[Chang et~al.(2022)Chang, Zhang, Jiang, Liu, and
  Freeman]{MARGEN:MASKGIT}
Huiwen Chang, Han Zhang, Lu~Jiang, Ce~Liu, and William~T Freeman.
\newblock Maskgit: Masked generative image transformer.
\newblock In \emph{Proceedings of the IEEE/CVF Conference on Computer Vision
  and Pattern Recognition}, pp.\  11315--11325, 2022.

\bibitem[Chang et~al.(2023)Chang, Zhang, Barber, Maschinot, Lezama, Jiang,
  Yang, Murphy, Freeman, Rubinstein, et~al.]{MARGEN:MUSE}
Huiwen Chang, Han Zhang, Jarred Barber, AJ~Maschinot, Jose Lezama, Lu~Jiang,
  Ming-Hsuan Yang, Kevin Murphy, William~T Freeman, Michael Rubinstein, et~al.
\newblock Muse: Text-to-image generation via masked generative transformers.
\newblock \emph{arXiv preprint arXiv:2301.00704}, 2023.

\bibitem[Chen et~al.(2024{\natexlab{a}})Chen, Zhang, Cun, Xia, Wang, Weng, and
  Shan]{VIDEODPM:VIDEOCRAFTER2}
Haoxin Chen, Yong Zhang, Xiaodong Cun, Menghan Xia, Xintao Wang, Chao Weng, and
  Ying Shan.
\newblock Videocrafter2: Overcoming data limitations for high-quality video
  diffusion models, 2024{\natexlab{a}}.

\bibitem[Chen et~al.(2023)Chen, Yu, Ge, Yao, Xie, Wu, Wang, Kwok, Luo, Lu, and
  Li]{DPM:PIXART}
Junsong Chen, Jincheng Yu, Chongjian Ge, Lewei Yao, Enze Xie, Yue Wu, Zhongdao
  Wang, James Kwok, Ping Luo, Huchuan Lu, and Zhenguo Li.
\newblock Pixart-$\alpha$: Fast training of diffusion transformer for
  photorealistic text-to-image synthesis, 2023.

\bibitem[Chen et~al.(2024{\natexlab{b}})Chen, Ge, Xie, Wu, Yao, Ren, Wang, Luo,
  Lu, and Li]{chen2024pixart}
Junsong Chen, Chongjian Ge, Enze Xie, Yue Wu, Lewei Yao, Xiaozhe Ren, Zhongdao
  Wang, Ping Luo, Huchuan Lu, and Zhenguo Li.
\newblock Pixart-sigma: Weak-to-strong training of diffusion transformer for 4k
  text-to-image generation.
\newblock In \emph{European Conference on Computer Vision}, pp.\  74--91.
  Springer, 2024{\natexlab{b}}.

\bibitem[Chen et~al.(2024{\natexlab{c}})Chen, Siarohin, Menapace, Deyneka,
  Chao, Jeon, Fang, Lee, Ren, Yang, et~al.]{DATA:PANDA}
Tsai-Shien Chen, Aliaksandr Siarohin, Willi Menapace, Ekaterina Deyneka,
  Hsiang-wei Chao, Byung~Eun Jeon, Yuwei Fang, Hsin-Ying Lee, Jian Ren,
  Ming-Hsuan Yang, et~al.
\newblock Panda-70m: Captioning 70m videos with multiple cross-modality
  teachers.
\newblock In \emph{Proceedings of the IEEE/CVF Conference on Computer Vision
  and Pattern Recognition}, pp.\  13320--13331, 2024{\natexlab{c}}.

\bibitem[Diao et~al.(2024)Diao, Cui, Li, Wang, Lu, and Wang]{VLM:EVE}
Haiwen Diao, Yufeng Cui, Xiaotong Li, Yueze Wang, Huchuan Lu, and Xinlong Wang.
\newblock Unveiling encoder-free vision-language models.
\newblock \emph{arXiv preprint arXiv:2406.11832}, 2024.

\bibitem[Ding et~al.(2021)Ding, Yang, Hong, Zheng, Zhou, Yin, Lin, Zou, Shao,
  Yang, et~al.]{MARGEN:COGVIEW1}
Ming Ding, Zhuoyi Yang, Wenyi Hong, Wendi Zheng, Chang Zhou, Da~Yin, Junyang
  Lin, Xu~Zou, Zhou Shao, Hongxia Yang, et~al.
\newblock Cogview: Mastering text-to-image generation via transformers.
\newblock \emph{Advances in neural information processing systems},
  34:\penalty0 19822--19835, 2021.

\bibitem[Ding et~al.(2022)Ding, Zheng, Hong, and Tang]{MARGEN:COGVIEW2}
Ming Ding, Wendi Zheng, Wenyi Hong, and Jie Tang.
\newblock Cogview2: Faster and better text-to-image generation via hierarchical
  transformers.
\newblock \emph{Advances in Neural Information Processing Systems},
  35:\penalty0 16890--16902, 2022.

\bibitem[Dosovitskiy et~al.(2021)Dosovitskiy, Beyer, Kolesnikov, Weissenborn,
  Zhai, Unterthiner, Dehghani, Minderer, Heigold, Gelly, Uszkoreit, and
  Houlsby]{TransF:ViT}
Alexey Dosovitskiy, Lucas Beyer, Alexander Kolesnikov, Dirk Weissenborn,
  Xiaohua Zhai, Thomas Unterthiner, Mostafa Dehghani, Matthias Minderer, Georg
  Heigold, Sylvain Gelly, Jakob Uszkoreit, and Neil Houlsby.
\newblock An image is worth 16x16 words: Transformers for image recognition at
  scale.
\newblock In \emph{ICLR}, 2021.

\bibitem[Elfwing et~al.(2018)Elfwing, Uchibe, and Doya]{SETUP:SILU}
Stefan Elfwing, Eiji Uchibe, and Kenji Doya.
\newblock Sigmoid-weighted linear units for neural network function
  approximation in reinforcement learning.
\newblock \emph{Neural networks}, 107:\penalty0 3--11, 2018.

\bibitem[Esser et~al.(2021)Esser, Rombach, and Ommer]{CAUGEN:VQGAN}
Patrick Esser, Robin Rombach, and Bjorn Ommer.
\newblock Taming transformers for high-resolution image synthesis.
\newblock In \emph{Proceedings of the IEEE/CVF conference on computer vision
  and pattern recognition}, pp.\  12873--12883, 2021.

\bibitem[Esser et~al.(2023)Esser, Chiu, Atighehchian, Granskog, and
  Germanidis]{VIDEODPM:StrDPM}
Patrick Esser, Johnathan Chiu, Parmida Atighehchian, Jonathan Granskog, and
  Anastasis Germanidis.
\newblock Structure and content-guided video synthesis with diffusion models.
\newblock In \emph{Proceedings of the IEEE/CVF International Conference on
  Computer Vision}, pp.\  7346--7356, 2023.

\bibitem[Esser et~al.(2024{\natexlab{a}})Esser, Kulal, Blattmann, Entezari,
  M{\"u}ller, Saini, Levi, Lorenz, Sauer, Boesel, et~al.]{DPM:SD3}
Patrick Esser, Sumith Kulal, Andreas Blattmann, Rahim Entezari, Jonas
  M{\"u}ller, Harry Saini, Yam Levi, Dominik Lorenz, Axel Sauer, Frederic
  Boesel, et~al.
\newblock Scaling rectified flow transformers for high-resolution image
  synthesis.
\newblock In \emph{Forty-first International Conference on Machine Learning},
  2024{\natexlab{a}}.

\bibitem[Esser et~al.(2024{\natexlab{b}})Esser, Kulal, Blattmann, Entezari,
  M{\"u}ller, Saini, Levi, Lorenz, Sauer, Boesel, et~al.]{SD3}
Patrick Esser, Sumith Kulal, Andreas Blattmann, Rahim Entezari, Jonas
  M{\"u}ller, Harry Saini, Yam Levi, Dominik Lorenz, Axel Sauer, Frederic
  Boesel, et~al.
\newblock Scaling rectified flow transformers for high-resolution image
  synthesis.
\newblock In \emph{Forty-first International Conference on Machine Learning},
  2024{\natexlab{b}}.

\bibitem[Gadre et~al.(2024)Gadre, Ilharco, Fang, Hayase, Smyrnis, Nguyen,
  Marten, Wortsman, Ghosh, Zhang, et~al.]{DATA:DATACOMP}
Samir~Yitzhak Gadre, Gabriel Ilharco, Alex Fang, Jonathan Hayase, Georgios
  Smyrnis, Thao Nguyen, Ryan Marten, Mitchell Wortsman, Dhruba Ghosh, Jieyu
  Zhang, et~al.
\newblock Datacomp: In search of the next generation of multimodal datasets.
\newblock \emph{Advances in Neural Information Processing Systems}, 2024.

\bibitem[Ghosh et~al.(2024)Ghosh, Hajishirzi, and Schmidt]{ghosh2024geneval}
Dhruba Ghosh, Hannaneh Hajishirzi, and Ludwig Schmidt.
\newblock Geneval: An object-focused framework for evaluating text-to-image
  alignment.
\newblock \emph{Advances in Neural Information Processing Systems}, 36, 2024.

\bibitem[Goyal(2017)]{SETUP:TRAINONEHPU}
P~Goyal.
\newblock Accurate, large minibatch sg d: training imagenet in 1 hour.
\newblock \emph{arXiv preprint arXiv:1706.02677}, 2017.

\bibitem[Guo et~al.(2023)Guo, Yang, Rao, Liang, Wang, Qiao, Agrawala, Lin, and
  Dai]{VIDEODPM:ANIMATEDPM}
Yuwei Guo, Ceyuan Yang, Anyi Rao, Zhengyang Liang, Yaohui Wang, Yu~Qiao,
  Maneesh Agrawala, Dahua Lin, and Bo~Dai.
\newblock Animatediff: Animate your personalized text-to-image diffusion models
  without specific tuning.
\newblock \emph{arXiv preprint arXiv:2307.04725}, 2023.

\bibitem[Guo et~al.(2024)Guo, Yang, Rao, Liang, Wang, Qiao, Agrawala, Lin, and
  Dai]{VIDEODPM:AnimateDiff}
Yuwei Guo, Ceyuan Yang, Anyi Rao, Zhengyang Liang, Yaohui Wang, Yu~Qiao,
  Maneesh Agrawala, Dahua Lin, and Bo~Dai.
\newblock Animatediff: Animate your personalized text-to-image diffusion models
  without specific tuning.
\newblock \emph{International Conference on Learning Representations}, 2024.

\bibitem[He et~al.(2022)He, Chen, Xie, Li, Doll{\'{a}}r, and
  Girshick]{TransF:MAE}
Kaiming He, Xinlei Chen, Saining Xie, Yanghao Li, Piotr Doll{\'{a}}r, and
  Ross~B. Girshick.
\newblock Masked autoencoders are scalable vision learners.
\newblock In \emph{CVPR}, pp.\  15979--15988, 2022.

\bibitem[Ho \& Salimans(2022)Ho and Salimans]{ho2022classifier}
Jonathan Ho and Tim Salimans.
\newblock Classifier-free diffusion guidance.
\newblock \emph{arXiv preprint arXiv:2207.12598}, 2022.

\bibitem[Ho et~al.(2020)Ho, Jain, and Abbeel]{DPM:DDPM}
Jonathan Ho, Ajay Jain, and Pieter Abbeel.
\newblock Denoising diffusion probabilistic models.
\newblock \emph{Advances in neural information processing systems},
  33:\penalty0 6840--6851, 2020.

\bibitem[Hong et~al.(2022)Hong, Ding, Zheng, Liu, and Tang]{MARGEN:COGVIDEO}
Wenyi Hong, Ming Ding, Wendi Zheng, Xinghan Liu, and Jie Tang.
\newblock Cogvideo: Large-scale pretraining for text-to-video generation via
  transformers.
\newblock \emph{arXiv preprint arXiv:2205.15868}, 2022.

\bibitem[Hoogeboom et~al.(2023)Hoogeboom, Heek, and Salimans]{DPM:SIMDPM}
Emiel Hoogeboom, Jonathan Heek, and Tim Salimans.
\newblock simple diffusion: End-to-end diffusion for high resolution images.
\newblock In \emph{International Conference on Machine Learning}, pp.\
  13213--13232. PMLR, 2023.

\bibitem[Hu et~al.(2024)Hu, Wang, Fang, Fu, Cheng, and Yu]{hu2024ella}
Xiwei Hu, Rui Wang, Yixiao Fang, Bin Fu, Pei Cheng, and Gang Yu.
\newblock Ella: Equip diffusion models with llm for enhanced semantic
  alignment.
\newblock \emph{arXiv preprint arXiv:2403.05135}, 2024.

\bibitem[Huang et~al.(2016)Huang, Sun, Liu, Sedra, and
  Weinberger]{SETUP:STODEPTH}
Gao Huang, Yu~Sun, Zhuang Liu, Daniel Sedra, and Kilian~Q Weinberger.
\newblock Deep networks with stochastic depth.
\newblock In \emph{Computer Vision--ECCV 2016: 14th European Conference,
  Amsterdam, The Netherlands, October 11--14, 2016, Proceedings, Part IV 14},
  pp.\  646--661. Springer, 2016.

\bibitem[Huang et~al.(2023)Huang, Sun, Xie, Li, and Liu]{SETUP:T2ICOMBENCH}
Kaiyi Huang, Kaiyue Sun, Enze Xie, Zhenguo Li, and Xihui Liu.
\newblock T2i-compbench: A comprehensive benchmark for open-world compositional
  text-to-image generation.
\newblock \emph{Advances in Neural Information Processing Systems}, 2023.

\bibitem[Huang \& Belongie(2017{\natexlab{a}})Huang and Belongie]{SETUP:ADLN}
Xun Huang and Serge Belongie.
\newblock Arbitrary style transfer in real-time with adaptive instance
  normalization.
\newblock In \emph{Proceedings of the IEEE international conference on computer
  vision}, pp.\  1501--1510, 2017{\natexlab{a}}.

\bibitem[Huang \& Belongie(2017{\natexlab{b}})Huang and
  Belongie]{huang2017arbitrary}
Xun Huang and Serge Belongie.
\newblock Arbitrary style transfer in real-time with adaptive instance
  normalization.
\newblock In \emph{Proceedings of the IEEE international conference on computer
  vision}, pp.\  1501--1510, 2017{\natexlab{b}}.

\bibitem[Huang et~al.(2024)Huang, He, Yu, Zhang, Si, Jiang, Zhang, Wu, Jin,
  Chanpaisit, et~al.]{SETUP:VBENCH}
Ziqi Huang, Yinan He, Jiashuo Yu, Fan Zhang, Chenyang Si, Yuming Jiang, Yuanhan
  Zhang, Tianxing Wu, Qingyang Jin, Nattapol Chanpaisit, et~al.
\newblock Vbench: Comprehensive benchmark suite for video generative models.
\newblock In \emph{Proceedings of the IEEE/CVF Conference on Computer Vision
  and Pattern Recognition}, pp.\  21807--21818, 2024.

\bibitem[Javaheripi et~al.(2023)Javaheripi, Bubeck, Abdin, Aneja, Bubeck,
  Mendes, Chen, Del~Giorno, Eldan, Gopi, et~al.]{LLM:PHI2}
Mojan Javaheripi, S{\'e}bastien Bubeck, Marah Abdin, Jyoti Aneja, Sebastien
  Bubeck, Caio C{\'e}sar~Teodoro Mendes, Weizhu Chen, Allie Del~Giorno, Ronen
  Eldan, Sivakanth Gopi, et~al.
\newblock Phi-2: The surprising power of small language models.
\newblock \emph{Microsoft Research Blog}, 2023.

\bibitem[Kalchbrenner et~al.(2017)Kalchbrenner, Oord, Simonyan, Danihelka,
  Vinyals, Graves, and Kavukcuoglu]{CAUGEN:VideoPixel}
Nal Kalchbrenner, A{\"a}ron Oord, Karen Simonyan, Ivo Danihelka, Oriol Vinyals,
  Alex Graves, and Koray Kavukcuoglu.
\newblock Video pixel networks.
\newblock In \emph{International Conference on Machine Learning}, pp.\
  1771--1779. PMLR, 2017.

\bibitem[Karras et~al.(2019)Karras, Laine, and Aila]{karras2019style}
Tero Karras, Samuli Laine, and Timo Aila.
\newblock A style-based generator architecture for generative adversarial
  networks.
\newblock In \emph{Proceedings of the IEEE/CVF conference on computer vision
  and pattern recognition}, pp.\  4401--4410, 2019.

\bibitem[Kondratyuk et~al.(2023)Kondratyuk, Yu, Gu, Lezama, Huang, Hornung,
  Adam, Akbari, Alon, Birodkar, et~al.]{CAUGEN:VIDEOPOET}
Dan Kondratyuk, Lijun Yu, Xiuye Gu, Jos{\'e} Lezama, Jonathan Huang, Rachel
  Hornung, Hartwig Adam, Hassan Akbari, Yair Alon, Vighnesh Birodkar, et~al.
\newblock Videopoet: A large language model for zero-shot video generation.
\newblock \emph{arXiv preprint arXiv:2312.14125}, 2023.

\bibitem[Kuaishou(2024)]{DPM:Kling}
Kuaishou.
\newblock Kling ai, 2024.
\newblock URL \url{https://klingai.com/}.

\bibitem[Lei~Ba et~al.(2016)Lei~Ba, Kiros, and Hinton]{SETUP:LN}
Jimmy Lei~Ba, Jamie~Ryan Kiros, and Geoffrey~E Hinton.
\newblock Layer normalization.
\newblock \emph{ArXiv e-prints}, pp.\  arXiv--1607, 2016.

\bibitem[Li et~al.(2024{\natexlab{a}})Li, Kamko, Akhgari, Sabet, Xu, and
  Doshi]{li2024playground}
Daiqing Li, Aleks Kamko, Ehsan Akhgari, Ali Sabet, Linmiao Xu, and Suhail
  Doshi.
\newblock Playground v2. 5: Three insights towards enhancing aesthetic quality
  in text-to-image generation.
\newblock \emph{arXiv preprint arXiv:2402.17245}, 2024{\natexlab{a}}.

\bibitem[Li et~al.(2024{\natexlab{b}})Li, Feng, Fu, Wang, Basu, Chen, and
  Wang]{VIDEODPM:T2VTurbo}
Jiachen Li, Weixi Feng, Tsu-Jui Fu, Xinyi Wang, Sugato Basu, Wenhu Chen, and
  William~Yang Wang.
\newblock T2v-turbo: Breaking the quality bottleneck of video consistency model
  with mixed reward feedback.
\newblock \emph{arXiv preprint arXiv:2405.18750}, 2024{\natexlab{b}}.

\bibitem[Li et~al.(2024{\natexlab{c}})Li, Tian, Li, Deng, and He]{MARGEN:MAR}
Tianhong Li, Yonglong Tian, He~Li, Mingyang Deng, and Kaiming He.
\newblock Autoregressive image generation without vector quantization.
\newblock \emph{arXiv preprint arXiv:2406.11838}, 2024{\natexlab{c}}.

\bibitem[Li et~al.(2024{\natexlab{d}})Li, Zhang, Lin, Xiong, Long, Deng, Zhang,
  Liu, Huang, Xiao, et~al.]{li2024hunyuan}
Zhimin Li, Jianwei Zhang, Qin Lin, Jiangfeng Xiong, Yanxin Long, Xinchi Deng,
  Yingfang Zhang, Xingchao Liu, Minbin Huang, Zedong Xiao, et~al.
\newblock Hunyuan-dit: A powerful multi-resolution diffusion transformer with
  fine-grained chinese understanding.
\newblock \emph{arXiv preprint arXiv:2405.08748}, 2024{\natexlab{d}}.

\bibitem[Liew et~al.(2023)Liew, Yan, Zhang, Xu, and Feng]{VIDEODPM:MAGICEDIT}
Jun~Hao Liew, Hanshu Yan, Jianfeng Zhang, Zhongcong Xu, and Jiashi Feng.
\newblock Magicedit: High-fidelity and temporally coherent video editing.
\newblock \emph{arXiv preprint arXiv:2308.14749}, 2023.

\bibitem[Lin et~al.(2024)Lin, Ge, Cheng, Li, Zhu, Wang, He, Ye, Yuan, Chen,
  et~al.]{VIDEODPM:OPENSORAPLAN}
Bin Lin, Yunyang Ge, Xinhua Cheng, Zongjian Li, Bin Zhu, Shaodong Wang, Xianyi
  He, Yang Ye, Shenghai Yuan, Liuhan Chen, et~al.
\newblock Open-sora plan: Open-source large video generation model, 2024.

\bibitem[Liu et~al.(2024)Liu, Akhgari, Visheratin, Kamko, Xu, Shrirao, Lambert,
  Souza, Doshi, and Li]{liu2024playground}
Bingchen Liu, Ehsan Akhgari, Alexander Visheratin, Aleks Kamko, Linmiao Xu,
  Shivam Shrirao, Chase Lambert, Joao Souza, Suhail Doshi, and Daiqing Li.
\newblock Playground v3: Improving text-to-image alignment with deep-fusion
  large language models.
\newblock \emph{arXiv preprint arXiv:2409.10695}, 2024.

\bibitem[Liu et~al.(2022)Liu, Hu, Lin, Yao, Xie, Wei, Ning, Cao, Zhang, Dong,
  et~al.]{TransF:SwinTransformerV2}
Ze~Liu, Han Hu, Yutong Lin, Zhuliang Yao, Zhenda Xie, Yixuan Wei, Jia Ning, Yue
  Cao, Zheng Zhang, Li~Dong, et~al.
\newblock Swin transformer v2: Scaling up capacity and resolution.
\newblock In \emph{Proceedings of the IEEE/CVF conference on computer vision
  and pattern recognition}, pp.\  12009--12019, 2022.

\bibitem[Loshchilov et~al.(2017)Loshchilov, Hutter, et~al.]{SETUP:ADAMW}
Ilya Loshchilov, Frank Hutter, et~al.
\newblock Fixing weight decay regularization in adam.
\newblock \emph{arXiv preprint arXiv:1711.05101}, 2017.

\bibitem[Meng et~al.(2021)Meng, He, Song, Song, Wu, Zhu, and Ermon]{DPM:SDEDIT}
Chenlin Meng, Yutong He, Yang Song, Jiaming Song, Jiajun Wu, Jun-Yan Zhu, and
  Stefano Ermon.
\newblock Sdedit: Guided image synthesis and editing with stochastic
  differential equations.
\newblock \emph{arXiv preprint arXiv:2108.01073}, 2021.

\bibitem[Nash et~al.(2022)Nash, Carreira, Walker, Barr, Jaegle, Malinowski, and
  Battaglia]{CAUGEN:TRANSFRMSER}
Charlie Nash, Joao Carreira, Jacob Walker, Iain Barr, Andrew Jaegle, Mateusz
  Malinowski, and Peter Battaglia.
\newblock Transframer: Arbitrary frame prediction with generative models.
\newblock \emph{arXiv preprint arXiv:2203.09494}, 2022.

\bibitem[Nichol et~al.(2021)Nichol, Dhariwal, Ramesh, Shyam, Mishkin, McGrew,
  Sutskever, and Chen]{DPM:GLIDE}
Alex Nichol, Prafulla Dhariwal, Aditya Ramesh, Pranav Shyam, Pamela Mishkin,
  Bob McGrew, Ilya Sutskever, and Mark Chen.
\newblock Glide: Towards photorealistic image generation and editing with
  text-guided diffusion models.
\newblock \emph{arXiv preprint arXiv:2112.10741}, 2021.

\bibitem[Nichol \& Dhariwal(2021)Nichol and Dhariwal]{DPM:IDDPM}
Alexander~Quinn Nichol and Prafulla Dhariwal.
\newblock Improved denoising diffusion probabilistic models.
\newblock In \emph{International conference on machine learning}, pp.\
  8162--8171. PMLR, 2021.

\bibitem[Peebles \& Xie(2023)Peebles and Xie]{peebles2023scalable}
William Peebles and Saining Xie.
\newblock Scalable diffusion models with transformers.
\newblock In \emph{Proceedings of the IEEE/CVF International Conference on
  Computer Vision}, pp.\  4195--4205, 2023.

\bibitem[Perez et~al.(2018)Perez, Strub, De~Vries, Dumoulin, and
  Courville]{perez2018film}
Ethan Perez, Florian Strub, Harm De~Vries, Vincent Dumoulin, and Aaron
  Courville.
\newblock Film: Visual reasoning with a general conditioning layer.
\newblock In \emph{Proceedings of the AAAI conference on artificial
  intelligence}, volume~32, 2018.

\bibitem[PexelsTeam(2014)]{DATA:Pexels}
PexelsTeam.
\newblock Pexels, royalty-free stock footage website, 2014.

\bibitem[Podell et~al.(2023)Podell, English, Lacey, Blattmann, Dockhorn,
  M{\"u}ller, Penna, and Rombach]{DPM:SDXL}
Dustin Podell, Zion English, Kyle Lacey, Andreas Blattmann, Tim Dockhorn, Jonas
  M{\"u}ller, Joe Penna, and Robin Rombach.
\newblock Sdxl: Improving latent diffusion models for high-resolution image
  synthesis.
\newblock \emph{arXiv preprint arXiv:2307.01952}, 2023.

\bibitem[Radford(2018)]{TransF:GPT-1}
Alec Radford.
\newblock Improving language understanding by generative pre-training, 2018.

\bibitem[Radford et~al.(2019)Radford, Wu, Child, Luan, Amodei, Sutskever,
  et~al.]{TransF:GPT-2}
Alec Radford, Jeffrey Wu, Rewon Child, David Luan, Dario Amodei, Ilya
  Sutskever, et~al.
\newblock Language models are unsupervised multitask learners.
\newblock \emph{OpenAI blog}, 1\penalty0 (8):\penalty0 9, 2019.

\bibitem[Ramesh et~al.(2021)Ramesh, Pavlov, Goh, Gray, Voss, Radford, Chen, and
  Sutskever]{MARGEN:DALLE1}
Aditya Ramesh, Mikhail Pavlov, Gabriel Goh, Scott Gray, Chelsea Voss, Alec
  Radford, Mark Chen, and Ilya Sutskever.
\newblock Zero-shot text-to-image generation.
\newblock In \emph{International conference on machine learning}, pp.\
  8821--8831. Pmlr, 2021.

\bibitem[Ramesh et~al.(2022)Ramesh, Dhariwal, Nichol, Chu, and
  Chen]{DPM:DALLE2}
Aditya Ramesh, Prafulla Dhariwal, Alex Nichol, Casey Chu, and Mark Chen.
\newblock Hierarchical text-conditional image generation with clip latents.
\newblock \emph{arXiv preprint arXiv:2204.06125}, 1\penalty0 (2):\penalty0 3,
  2022.

\bibitem[Reed et~al.(2017)Reed, Oord, Kalchbrenner, Colmenarejo, Wang, Chen,
  Belov, and Freitas]{CAUGEN:PARALLEL}
Scott Reed, A{\"a}ron Oord, Nal Kalchbrenner, Sergio~G{\'o}mez Colmenarejo,
  Ziyu Wang, Yutian Chen, Dan Belov, and Nando Freitas.
\newblock Parallel multiscale autoregressive density estimation.
\newblock In \emph{International conference on machine learning}, pp.\
  2912--2921. PMLR, 2017.

\bibitem[Rombach et~al.(2022{\natexlab{a}})Rombach, Blattmann, Lorenz, Esser,
  and Ommer]{DPM:LDM}
Robin Rombach, Andreas Blattmann, Dominik Lorenz, Patrick Esser, and Bj{\"o}rn
  Ommer.
\newblock High-resolution image synthesis with latent diffusion models.
\newblock In \emph{Proceedings of the IEEE/CVF conference on computer vision
  and pattern recognition}, pp.\  10684--10695, 2022{\natexlab{a}}.

\bibitem[Rombach et~al.(2022{\natexlab{b}})Rombach, Blattmann, Lorenz, Esser,
  and Ommer]{rombach2022high}
Robin Rombach, Andreas Blattmann, Dominik Lorenz, Patrick Esser, and Bj{\"o}rn
  Ommer.
\newblock High-resolution image synthesis with latent diffusion models.
\newblock In \emph{Proceedings of the IEEE/CVF conference on computer vision
  and pattern recognition}, pp.\  10684--10695, 2022{\natexlab{b}}.

\bibitem[Runway(2023)]{VIDEODPM:RUNWAYGEN2}
Runway.
\newblock Gen-2: Generate novel videos with text, images or video clips, 2023.
\newblock URL \url{https://runwayml.com/research/gen-2}.

\bibitem[Runway(2024)]{VIDEODPM:Gen3}
Runway.
\newblock Gen-3 alpha: A new frontier for video generation, 2024.
\newblock URL \url{https://runwayml.com/research/introducing-gen-3-alpha}.

\bibitem[Schuhmann et~al.(2022)Schuhmann, Beaumont, Vencu, Gordon, Wightman,
  Cherti, Coombes, Katta, Mullis, Wortsman, et~al.]{Datasets:Laion-5b}
Christoph Schuhmann, Romain Beaumont, Richard Vencu, Cade Gordon, Ross
  Wightman, Mehdi Cherti, Theo Coombes, Aarush Katta, Clayton Mullis, Mitchell
  Wortsman, et~al.
\newblock Laion-5b: An open large-scale dataset for training next generation
  image-text models.
\newblock \emph{NeurIPS}, 35:\penalty0 25278--25294, 2022.

\bibitem[Song et~al.(2020)Song, Sohl-Dickstein, Kingma, Kumar, Ermon, and
  Poole]{DPM:SMLM}
Yang Song, Jascha Sohl-Dickstein, Diederik~P Kingma, Abhishek Kumar, Stefano
  Ermon, and Ben Poole.
\newblock Score-based generative modeling through stochastic differential
  equations.
\newblock \emph{arXiv preprint arXiv:2011.13456}, 2020.

\bibitem[Sun et~al.(2024{\natexlab{a}})Sun, Pan, Ge, Li, Duan, Wu, Zhang, Zhou,
  Qin, Wang, et~al.]{DATA:JOURNEYDB}
Keqiang Sun, Junting Pan, Yuying Ge, Hao Li, Haodong Duan, Xiaoshi Wu, Renrui
  Zhang, Aojun Zhou, Zipeng Qin, Yi~Wang, et~al.
\newblock Journeydb: A benchmark for generative image understanding.
\newblock \emph{Advances in Neural Information Processing Systems},
  2024{\natexlab{a}}.

\bibitem[Sun et~al.(2024{\natexlab{b}})Sun, Jiang, Chen, Zhang, Peng, Luo, and
  Yuan]{CAUGEN:LLAMAGEN}
Peize Sun, Yi~Jiang, Shoufa Chen, Shilong Zhang, Bingyue Peng, Ping Luo, and
  Zehuan Yuan.
\newblock Autoregressive model beats diffusion: Llama for scalable image
  generation.
\newblock \emph{arXiv preprint arXiv:2406.06525}, 2024{\natexlab{b}}.

\bibitem[Sun et~al.(2023)Sun, Cui, Zhang, Zhang, Yu, Luo, Wang, Rao, Liu,
  Huang, and Wang]{VLM:EMUv2}
Quan Sun, Yufeng Cui, Xiaosong Zhang, Fan Zhang, Qiying Yu, Zhengxiong Luo,
  Yueze Wang, Yongming Rao, Jingjing Liu, Tiejun Huang, and Xinlong Wang.
\newblock Generative multimodal models are in-context learners.
\newblock \emph{arXiv: 2312.13286}, 2023.

\bibitem[Tian et~al.(2024)Tian, Jiang, Yuan, Peng, and Wang]{CAUGEN:VAR}
Keyu Tian, Yi~Jiang, Zehuan Yuan, Bingyue Peng, and Liwei Wang.
\newblock Visual autoregressive modeling: Scalable image generation via
  next-scale prediction.
\newblock \emph{arXiv preprint arXiv:2404.02905}, 2024.

\bibitem[Touvron et~al.(2023)Touvron, Martin, Stone, Albert, Almahairi, Babaei,
  Bashlykov, Batra, Bhargava, Bhosale, et~al.]{TransF:LLaMA2}
Hugo Touvron, Louis Martin, Kevin Stone, Peter Albert, Amjad Almahairi, Yasmine
  Babaei, Nikolay Bashlykov, Soumya Batra, Prajjwal Bhargava, Shruti Bhosale,
  et~al.
\newblock Llama 2: Open foundation and fine-tuned chat models.
\newblock \emph{arXiv: 2307.09288}, 2023.

\bibitem[UnsplashTeam(2020)]{DATA:Unsplash}
UnsplashTeam.
\newblock Unsplash dataset, 2020.

\bibitem[Van Den~Oord et~al.(2017)Van Den~Oord, Vinyals, et~al.]{CAUGEN:VQVAE}
Aaron Van Den~Oord, Oriol Vinyals, et~al.
\newblock Neural discrete representation learning.
\newblock \emph{Advances in neural information processing systems}, 30, 2017.

\bibitem[Vaswani et~al.(2017)Vaswani, Shazeer, Parmar, Uszkoreit, Jones, Gomez,
  Kaiser, and Polosukhin]{TransF:Transformer}
Ashish Vaswani, Noam Shazeer, Niki Parmar, Jakob Uszkoreit, Llion Jones,
  Aidan~N. Gomez, Lukasz Kaiser, and Illia Polosukhin.
\newblock Attention is all you need.
\newblock In \emph{NIPS}, pp.\  5998--6008, 2017.

\bibitem[Villegas et~al.(2022)Villegas, Babaeizadeh, Kindermans, Moraldo,
  Zhang, Saffar, Castro, Kunze, and Erhan]{MARGEN:PHENAKI}
Ruben Villegas, Mohammad Babaeizadeh, Pieter-Jan Kindermans, Hernan Moraldo,
  Han Zhang, Mohammad~Taghi Saffar, Santiago Castro, Julius Kunze, and Dumitru
  Erhan.
\newblock Phenaki: Variable length video generation from open domain textual
  descriptions.
\newblock In \emph{International Conference on Learning Representations}, 2022.

\bibitem[Wang et~al.(2024{\natexlab{a}})Wang, Zhang, Luo, Sun, Cui, Wang,
  Zhang, Wang, Li, Yu, et~al.]{EMU3}
Xinlong Wang, Xiaosong Zhang, Zhengxiong Luo, Quan Sun, Yufeng Cui, Jinsheng
  Wang, Fan Zhang, Yueze Wang, Zhen Li, Qiying Yu, et~al.
\newblock Emu3: Next-token prediction is all you need.
\newblock \emph{arXiv preprint arXiv:2409.18869}, 2024{\natexlab{a}}.

\bibitem[Wang et~al.(2023)Wang, Chen, Ma, Zhou, Huang, Wang, Yang, He, Yu,
  Yang, et~al.]{VIDEODPM:LAVIE}
Yaohui Wang, Xinyuan Chen, Xin Ma, Shangchen Zhou, Ziqi Huang, Yi~Wang, Ceyuan
  Yang, Yinan He, Jiashuo Yu, Peiqing Yang, et~al.
\newblock Lavie: High-quality video generation with cascaded latent diffusion
  models.
\newblock \emph{arXiv preprint arXiv:2309.15103}, 2023.

\bibitem[Wang et~al.(2024{\natexlab{b}})Wang, Xiong, Zhou, Lin, Zhao, Kang,
  Feng, and Liu]{CAUGEN:LOONG}
Yuqing Wang, Tianwei Xiong, Daquan Zhou, Zhijie Lin, Yang Zhao, Bingyi Kang,
  Jiashi Feng, and Xihui Liu.
\newblock Loong: Generating minute-level long videos with autoregressive
  language models.
\newblock \emph{arXiv preprint arXiv:2410.02757}, 2024{\natexlab{b}}.

\bibitem[Yan et~al.(2021)Yan, Zhang, Abbeel, and Srinivas]{CAUGEN:VIDEOGPT}
Wilson Yan, Yunzhi Zhang, Pieter Abbeel, and Aravind Srinivas.
\newblock Videogpt: Video generation using vq-vae and transformers.
\newblock \emph{arXiv preprint arXiv:2104.10157}, 2021.

\bibitem[Yang et~al.(2024)Yang, Teng, Zheng, Ding, Huang, Xu, Yang, Hong,
  Zhang, Feng, et~al.]{VIDEODPM:COGVIDEOX}
Zhuoyi Yang, Jiayan Teng, Wendi Zheng, Ming Ding, Shiyu Huang, Jiazheng Xu,
  Yuanming Yang, Wenyi Hong, Xiaohan Zhang, Guanyu Feng, et~al.
\newblock Cogvideox: Text-to-video diffusion models with an expert transformer.
\newblock \emph{arXiv preprint arXiv:2408.06072}, 2024.

\bibitem[Yu et~al.(2022)Yu, Xu, Koh, Luong, Baid, Wang, Vasudevan, Ku, Yang,
  Ayan, et~al.]{CAUGEN:PARTI}
Jiahui Yu, Yuanzhong Xu, Jing~Yu Koh, Thang Luong, Gunjan Baid, Zirui Wang,
  Vijay Vasudevan, Alexander Ku, Yinfei Yang, Burcu~Karagol Ayan, et~al.
\newblock Scaling autoregressive models for content-rich text-to-image
  generation.
\newblock \emph{arXiv preprint arXiv:2206.10789}, 2\penalty0 (3):\penalty0 5,
  2022.

\bibitem[Yu et~al.(2023)Yu, Cheng, Sohn, Lezama, Zhang, Chang, Hauptmann, Yang,
  Hao, Essa, et~al.]{MARGEN:MAGVIT}
Lijun Yu, Yong Cheng, Kihyuk Sohn, Jos{\'e} Lezama, Han Zhang, Huiwen Chang,
  Alexander~G Hauptmann, Ming-Hsuan Yang, Yuan Hao, Irfan Essa, et~al.
\newblock Magvit: Masked generative video transformer.
\newblock In \emph{Proceedings of the IEEE/CVF Conference on Computer Vision
  and Pattern Recognition}, pp.\  10459--10469, 2023.

\bibitem[Zhang et~al.(2023{\natexlab{a}})Zhang, Wu, Liu, Zhao, Ran, Gu, Gao,
  and Shou]{VIDEODPM:Show1}
David~Junhao Zhang, Jay~Zhangjie Wu, Jia-Wei Liu, Rui Zhao, Lingmin Ran, Yuchao
  Gu, Difei Gao, and Mike~Zheng Shou.
\newblock Show-1: Marrying pixel and latent diffusion models for text-to-video
  generation.
\newblock \emph{arXiv preprint arXiv:2309.15818}, 2023{\natexlab{a}}.

\bibitem[Zhang et~al.(2023{\natexlab{b}})Zhang, Rao, and
  Agrawala]{DPM:CONTROLNET}
Lvmin Zhang, Anyi Rao, and Maneesh Agrawala.
\newblock Adding conditional control to text-to-image diffusion models.
\newblock In \emph{Proceedings of the IEEE/CVF International Conference on
  Computer Vision}, pp.\  3836--3847, 2023{\natexlab{b}}.

\bibitem[Zhang et~al.(2018)Zhang, Isola, Efros, Shechtman, and
  Wang]{zhang2018unreasonable}
Richard Zhang, Phillip Isola, Alexei~A Efros, Eli Shechtman, and Oliver Wang.
\newblock The unreasonable effectiveness of deep features as a perceptual
  metric.
\newblock In \emph{Proceedings of the IEEE conference on computer vision and
  pattern recognition}, pp.\  586--595, 2018.

\bibitem[Zheng et~al.(2024)Zheng, Peng, Yang, Shen, Li, Liu, Zhou, Li, and
  You]{VIDEODPM:OPENSORA}
Zangwei Zheng, Xiangyu Peng, Tianji Yang, Chenhui Shen, Shenggui Li, Hongxin
  Liu, Yukun Zhou, Tianyi Li, and Yang You.
\newblock Open-sora: Democratizing efficient video production for all, March
  2024.
\newblock URL \url{https://github.com/hpcaitech/Open-Sora}.

\bibitem[Zhuo et~al.(2024)Zhuo, Du, Xiao, Li, Liu, Huang, Liu, Zhao, Wang, Ma,
  et~al.]{DPM:LUMINANEXT}
Le~Zhuo, Ruoyi Du, Han Xiao, Yangguang Li, Dongyang Liu, Rongjie Huang, Wenze
  Liu, Lirui Zhao, Fu-Yun Wang, Zhanyu Ma, et~al.
\newblock Lumina-next: Making lumina-t2x stronger and faster with next-dit.
\newblock \emph{arXiv preprint arXiv:2406.18583}, 2024.

\end{thebibliography}
\bibliographystyle{iclr2025_conference}

\clearpage
\appendix
\section*{Appendix}

We strictly publish our code and pretrained models to improve interpretability and assure reproducibility. Here, more implementation details and ablation experiments are organized as follows:
{\leftmargini=2em
\begin{itemize}
\item Architecture details of Scaling and Shift layer (Sec.~\ref{sec:supp_mixer})
\item Normalization configurations (Sec.~\ref{sec:supp_norm})
\item Video extrapolation evaluations (Sec.~\ref{sec:extrapolation})
\item Inference time analysis (Sec.~\ref{sec:infer})
\item Ablations on the impact of temporal autoregressive modeling (Sec.~\ref{sec:tam})
\item Comprehensive DPG-Bench evaluation results (Sec.~\ref{sec:dpg-bench})
\item More text-to-image visualizations (Sec.~\ref{sec:supp_t2i})
\item More text-to-video visualizations (Sec.~\ref{sec:supp_t2v})
\end{itemize}}

\section{Architecture details of Scaling and Shift layer}\label{sec:supp_mixer}

\begin{figure}[ht]
\begin{minipage}[t]{0.44\textwidth}
    \centering
    \includegraphics[width=0.99\linewidth,trim= 10 0 0 0,clip]{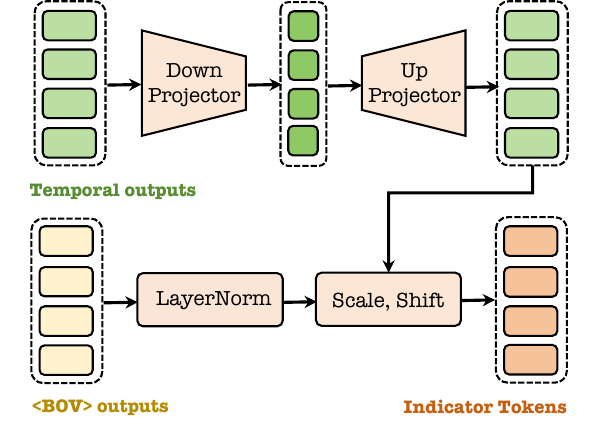}
    \caption{\textbf{Scaling and Shift layer.} We reformulate cross-frame motion changes by learning relative distribution variations within a unified space based on BOV tokens, rather than directly modeling the unreferenced distribution of the current frame.}
    \label{fig:supp_mixer}
\end{minipage}
\hfill%
\begin{minipage}[t]{0.53\textwidth}
    \centering
    \includegraphics[width=0.99\linewidth,trim= 0 0 10 0,clip]{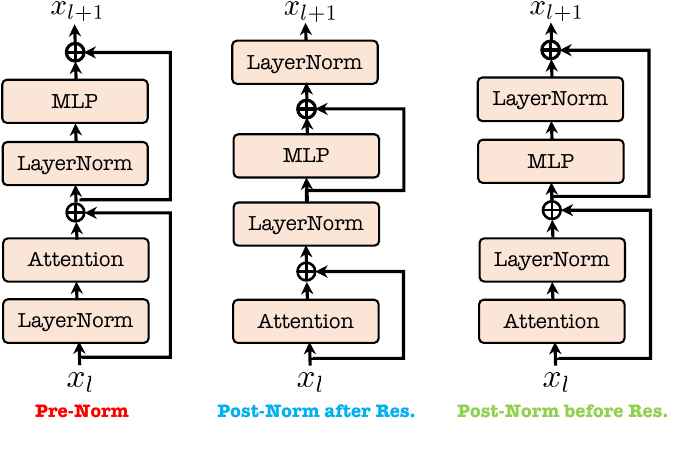}
    \caption{\textbf{Three normalization architectures.} We summarize various configurations including the pre-normalization layer (left), the post-normalization layer after residual addition (middle), and the post-normalization layer before residual addition (right). Here Post-Norm before Res is our standard design.}
    \label{fig:supp_norm}
\end{minipage}
\end{figure}

The Scaling and Shift Layer is implemented as an adaptive normalization layer, adopting the design initially proposed by FiLM (\cite{perez2018film}) and AdaIN (\cite{huang2017arbitrary}). While many previous methods have primarily utilized adaptive normalization for controllable image generation, such as in StyleGAN (\cite{karras2019style}), or for conditional modeling within Diffusion Transformers, like DiT (\cite{peebles2023scalable}), NOVA innovatively applies this technique to manage the cumulative inference errors in autoregressive video generation. We employ a two-layer MLP to optimize low-rank decomposition for motion changes, as shown in the Figure~\ref{fig:supp_mixer}. Specifically, we refer \texttt{AdaLayerNorm} and decompose the motion changes into mean and variance parameters, which are further used to apply the affine transformation on BOV embeddings.

\section{Normalization configurations}\label{sec:supp_norm}

NOVA employs an improved normalization configuration that can effectively control the numerical boundaries of the output embeddings of each Transformer block while also maintaining the identity transformation of residual connections. We illustrate the three common normalization configurations in Figure~\ref{fig:supp_norm}, and NOVA uses the \textit{post-normalization before residual addition} by default.

\section{video extrapolation evaluations}\label{sec:extrapolation}
Video extrapolation represents a significant challenge, being an out-of-domain generalization issue. To assess our model's performance, we curated a test set comprising 200 videos. For each video, the task involved generating subsequent frames from the initial frame and a textual prompt, effectively converting an image and text into a video sequence. We utilized LPIPS (\cite{zhang2018unreasonable}) and PSNR metrics to evaluate the video extrapolation capabilities of our model.

\begin{figure}[t]
\centering
\includegraphics[width=0.98\linewidth,trim= 0 0 0 0,clip]{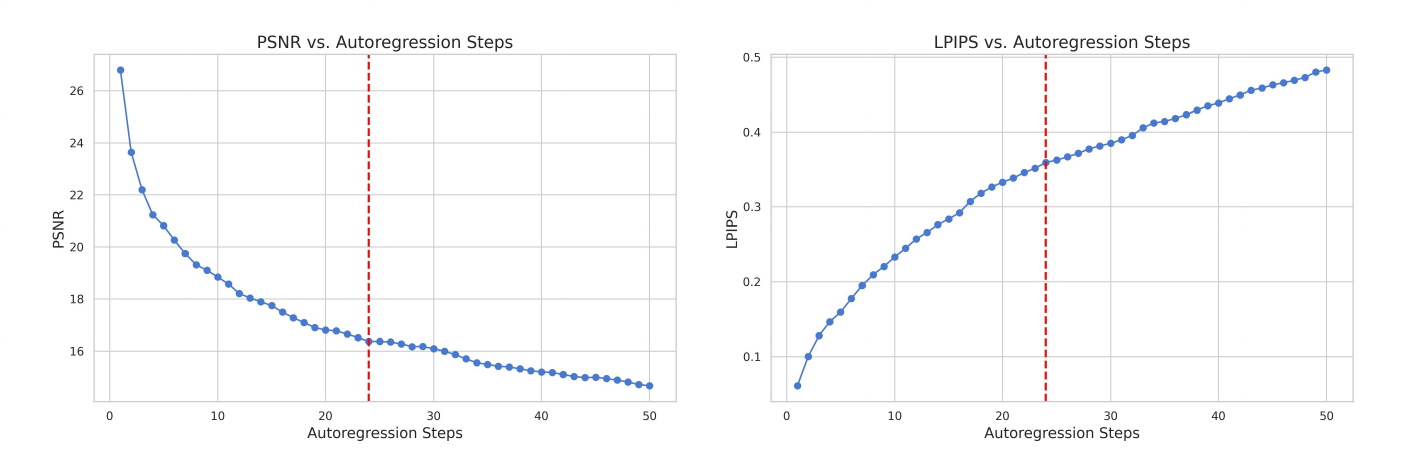}
\caption{\textbf{PSNR and LPIPS metrics over 50 autoregressive steps in video extrapolation. } Due to the 4$\times$ downsampling rate of VAE in temporal scale, each autoregressive step generates four frames. The vertical red line marks the point where the extrapolation reaches 3$\times$ training length.}
\label{fig:video_ex_psnr}
\end{figure}

\begin{figure}[t]
\centering
\includegraphics[width=0.98\linewidth,trim= 0 0 0 0,clip]{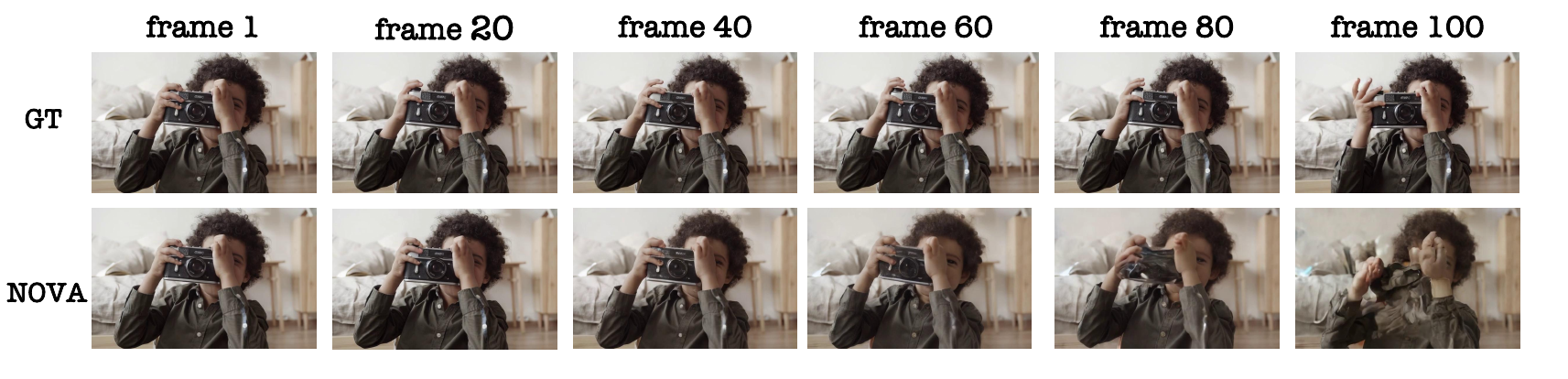}
\caption{\textbf{Visualization of video extrapolation.} Although the metrics indicate a decline, the generated frames still closely resemble the original video in content and overall image quality. Visualization suggests that the model can extrapolate up to 3$\times$ training length.}
\label{fig:video_ex_vis}
\end{figure}

During the extrapolation process, it was observed that the generated frames started to deviate from the ground truth after a few iterations. This is mainly due to the difficulty in accurately capturing video dynamics, causing minor discrepancies to accumulate. As a result, per-frame PSNR values decrease, while LPIPS scores increase over time (Figure~\ref{fig:video_ex_psnr}). Nevertheless, the generated frames exhibit a high degree of similarity to the original video in terms of both content and image quality in Figure~\ref{fig:video_ex_vis}. This highlights the robustness of our temporal autoregressive approach in video extrapolation.

\section{Inference time analysis}\label{sec:infer}
We report inference times on a single NVIDIA A100 GPU (40GB) with a batch size of 24 in Table~\ref{tab:time_consumption}. In each video, the temporal layers require only 0.03 seconds, compared to 11.97 seconds for the spatial layers, highlighting the exceptional efficiency of the temporal layers. While NOVA is already efficient in text-to-video generation, there is potential for further acceleration in the spatial layers.

\begin{table}[ht]
\centering
\caption{\textbf{Inference time analysis for different layers.}}
\label{tab:time_consumption}
\small
\resizebox{0.7\linewidth}{!}{
\begin{tabular}{c|c|c|c}
\toprule
Resolution & Temporal Layers Time & Spatial Layers Time & Total Time \\
\midrule
29$\times$768$\times$480 & 0.03s & 11.97s & 12s \\
\bottomrule
\end{tabular}·
}
\end{table}

\section{Ablations on the impact of Temporal Autoregressive Modeling}\label{sec:tam}
Under the same experimental settings, we evaluated VBench results both with and without using TAM (Temporal Autoregressive Modeling) to highlight its significance. Our findings are summarized as follows:
\textbf{(1) Efficient Motion Modeling:} We observed that the total score was marginally lower without TAM compared to NOVA (75.38 vs. 75.84), especially in the dynamic degree metric, which showed a more pronounced decline (11.38 vs. 23.27). We hypothesize that while bidirectional attention enhances model capacity, it requires more extensive data and longer training times to capture subtle motion changes compared to causal models.
\textbf{(2) Efficient Video Inference:} Thanks to the kv-cache technology and frame-by-frame autoregressive processing, NOVA's inference time is much faster compared to methods without TAM, with a greater speed advantage for longer videos.

\begin{table}[ht]
\centering
\caption{\textbf{Performance comparison on temporal autoregressive modeling.}}
\label{tab:tam_comparison}
\footnotesize
\resizebox{0.6\linewidth}{!}{
\begin{tabular}{l|c|c|c}
\toprule
Model & Total Score & Dynamic Degree & Infer Time \\
\midrule
NOVA & 75.84 & 23.27 & 12s \\
NOVA (w/o TAM) & 75.38 & 11.38 & 39s \\
\bottomrule
\end{tabular}
}
\end{table}

\section{Comprehensive DPG-Bench evaluation results}\label{sec:dpg-bench}

We provide detailed DPG-Bench scores in Table~\ref{tab:dpg}. While NOVA outperforms most models of comparable size and matches the overall score of state-of-the-art models, we observe that increasing the model scale results in marginal improvements and does not boost the text rendering performance. This limitation may be attributed to our reliance on extensive web datasets, such as LAION and DataComp. In future work, we plan to focus on improving the quality of text-to-image data.

\begin{table}[ht]
\centering
\caption{\textbf{Comparison with state-of-the-art models on DPG-Bench.}}
\label{tab:dpg}
\setlength{\tabcolsep}{2.5mm}
\resizebox{\linewidth}{!}{
\begin{tabular}{lcccccc}
\toprule 
Model & Overall & Global & Entity & Attribute & Relation & Other \\ 
\midrule
\multicolumn{7}{l}{\small\textit{Diffusion models}} \\
SD v1.5 (\cite{rombach2022high}) & 63.18 & 74.63 & 74.23 & 75.39 & 73.49 & 67.81 \\
PixArt-$\alpha$ (\cite{DPM:PIXART}) & 71.11 & 74.97 & 79.32 & 78.60 &  82.57 &  76.96 \\
PixArt-$\sigma$ (\cite{chen2024pixart}) & 80.54 & 86.89 & 82.89 & 88.94 & 86.59 & 87.68 \\
Lumina-Next (\cite{DPM:LUMINANEXT}) & 74.63 & 82.82 & 88.65 & 86.44 & 80.53 & 81.82 \\
SDXL (\cite{DPM:SDXL}) & 74.65 & 83.27 & 82.43 & 80.91 & 86.76 & 80.41 \\
Playground v2.5 (\cite{li2024playground}) & 75.47 & 83.06 & 82.59 & 81.20 & 84.08 & 83.50 \\
Hunyuan-DiT (\cite{li2024hunyuan}) & 78.87 & 84.59 & 80.59 & 88.01 & 74.36 & 86.41 \\
DALL-E3 (\cite{DPM:DALLE3}) & 83.50 & 90.97 & 89.61 & 88.39 & 90.58 & 89.83 \\
SD3 (\cite{SD3}) & 84.08 & 87.90 & 91.01 & 88.83 & 80.70 & 88.68 \\
Playground v3 (\cite{liu2024playground}) & 87.04 & 91.94 & 85.71 & 90.90 & 90.00 & 92.72 \\
\midrule
\multicolumn{7}{l}{\small\textit{Autoregressive models}} \\
Emu3-DPO (\cite{EMU3}) & 81.60 & 87.54 & 87.17 & 86.33 & 90.61 & 89.75 \\
\midrule
NOVA (0.3B) & 80.60 & 85.41 & 86.97 & 85.16 & 92.05 & 71.20 \\
NOVA (0.6B) & 82.25 & 87.65 & 87.65 & 85.62 & 90.90 & 74.80 \\
NOVA (1.4B) & 83.01 & 86.32 & 88.69 & 86.35 & 91.94 & 74.80 \\
\bottomrule
\end{tabular}}
\end{table}

\clearpage

\section{More text-to-image visualizations}\label{sec:supp_t2i}

\begin{figure}[ht]
\centering
\includegraphics[width=0.75\linewidth,trim= 0 0 0 0,clip]{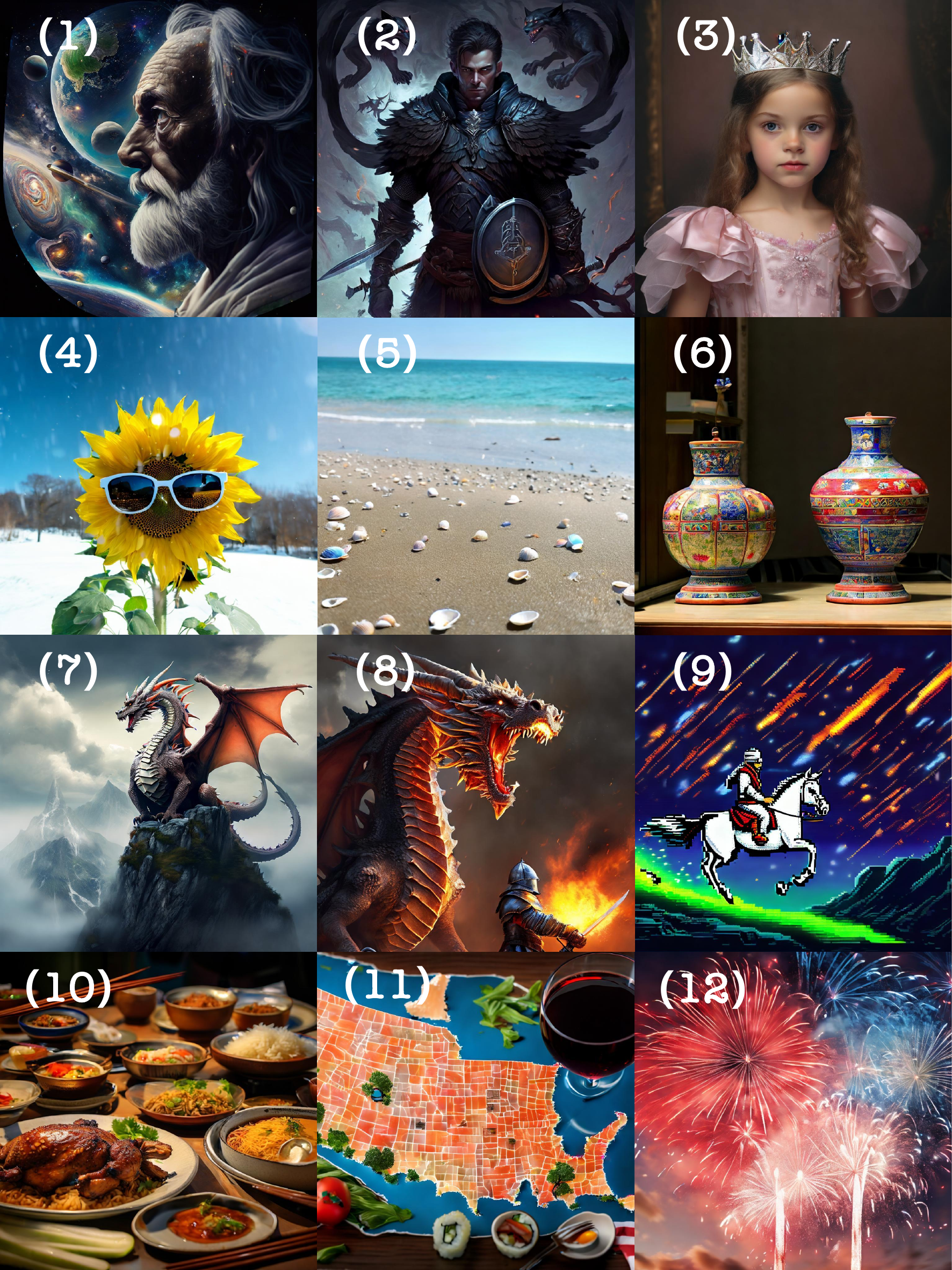}
\caption{\textbf{More text-to-image visualizations.} Text prompts are as follows: (1) ``In the foreground is the detailed, head-and-shoulders portrait of an elderly man with a long white beard...'', (2) ``a digital artwork of a fantasy warrior character. The character is male, depicted from the waist up, and appears to have a stern or serious facial expression...'', (3) ``a young girl wearing a tiara and frilly dress'', (4) ``A sunflower in sunglasses dances in the snow'', (5) ``A beach with no people'', (6) ``Two Ming vases on the table, the larger one is more colorful than the other'', (7) ``A dragon perched majestically on a craggy, smoke-wreathed mountain'', (8) ``a dragon breathing fire onto a knight'', (9) ``a pixel art style graphic with vibrant colors. It features a single rider on a horse, both depicted in mid-gallop to the left side of the frame...'', (10) ``A table full of food. There is a plate of chicken rice, a bowl of bak chor mee, and a bowl of laksa'', (11) ``A map of the United States made out sushi. It is on a table next to a glass of red wine'' and (12) ``beautiful fireworks in the sky with red, white and blue''.}
\label{fig:supp_t2i}
\end{figure}

We present more text-to-image samples in the Figure~\ref{fig:supp_t2i}. NOVA can generate images with a maximum resolution of 1024$\times$1024. Our model excels in the domain of text-to-image generation, producing a vast array of high-quality images that accurately reflect the textual descriptions provided. This capability not only spans a wide range of subjects, from realistic landscapes and portraits to imaginative and abstract concepts, but also maintains a high level of detail and aesthetic quality. 

\clearpage

\section{More text-to-video visualizations}\label{sec:supp_t2v}

\begin{figure}[ht]
\centering
\includegraphics[width=0.98\linewidth,trim= 0 225 0 0,clip]{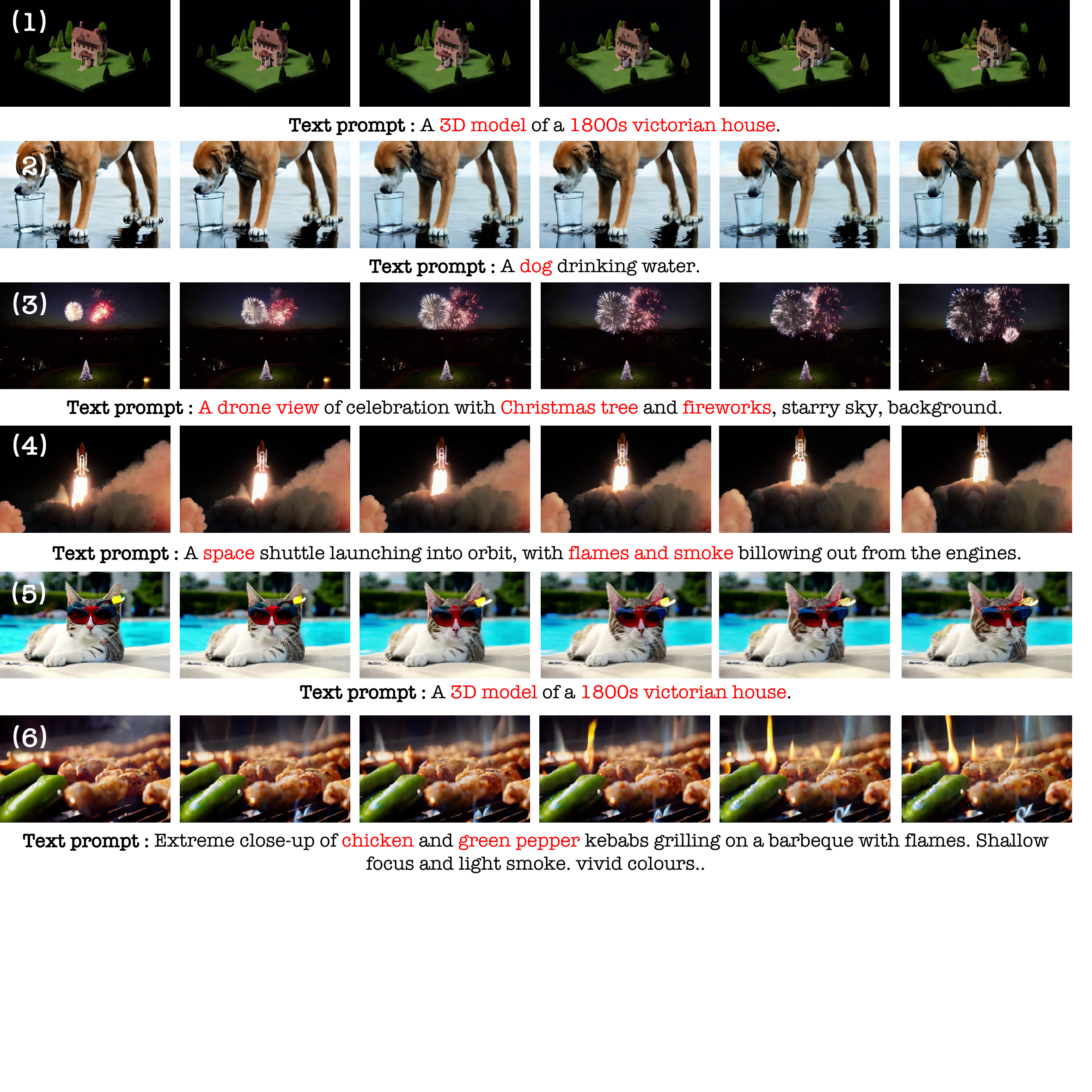} 
\caption{More text-to-video visualizations. Best viewed with zoom for enhanced detail.}
\label{fig:supp_t2v}
\end{figure}

We present more text-to-video samples generated by NOVA in the Figure~\ref{fig:supp_t2v}. NOVA can generate videos with a resolution of 33$\times$768$\times$480. Our model stands out in the field of text-to-video generation, capable of producing a substantial number of high-quality videos that vividly bring textual descriptions to life. From detailed storylines and character animations to realistic environmental settings and action scenes, our model demonstrates exceptional proficiency in generating content.

\clearpage
\end{document}